\documentclass[10pt,journal,compsoc]{IEEEtran}

\usepackage{graphicx}
\usepackage{amsmath, bm}
\usepackage{amssymb}
\usepackage{tabularx}
\usepackage[linesnumbered,vlined,ruled]{algorithm2e}
\usepackage[greek,english]{babel}
\usepackage{multicol,multirow,booktabs,mathabx}
\usepackage{makecell}
\usepackage{array}
\usepackage{float}
\usepackage{booktabs}
\usepackage[table]{xcolor}
\usepackage{pifont}
\usepackage{dblfloatfix}
\newcolumntype{P}[1]{>{\centering\arraybackslash}p{#1}}

\usepackage{arydshln}
\usepackage{setspace}
\usepackage[colorlinks=true, urlcolor=blue, linkcolor=red]{hyperref}
\usepackage{diagbox}

\ifCLASSOPTIONcompsoc
  \usepackage[nocompress]{cite}
\else
  \usepackage{cite}
\fi

\ifCLASSINFOpdf
\else
\fi

\usepackage{arydshln}

\begin{document}

\title{Visual Multi-Object Tracking with Re-Identification and Occlusion Handling using Labeled Random Finite Sets}

\author{Linh~Van~Ma, Tran~Thien~Dat~Nguyen, Changbeom~Shim, Du~Yong~Kim, Namkoo~Ha, Moongu~Jeon
\IEEEcompsocitemizethanks{\IEEEcompsocthanksitem Linh Van Ma and Moongu Jeon are with the School of Electrical Engineering and Computer Science at GIST, Gwangju, Korea (e-mail: \{linh.mavan, mgjeon\}@gist.ac.kr).
\IEEEcompsocthanksitem Tran Thien Dat Nguyen and Changbeom Shim are with the School of Electrical Engineering, Computing and Mathematical Sciences, Curtin University, Australia (e-mail: \{t.nguyen1, changbeom.shim\}@curtin.edu.au).
\IEEEcompsocthanksitem Du~Yong~Kim is with School of Engineering, RMIT University, Australia (e-mail: duyong.kim@rmit.edu.au).
\IEEEcompsocthanksitem Namkoo~Ha is with the Department of EO/IR Systems Research and Development, LIG Nex1, Korea (e-mail: namkoo.ha@lignex1.com).}}


\IEEEtitleabstractindextext{%
\begin{abstract}
This paper proposes an online visual multi-object tracking (MOT) algorithm that resolves object appearance-reappearance and occlusion. Our solution is based on the labeled random finite set (LRFS) filtering approach, which in principle, addresses disappearance, appearance, reappearance, and occlusion via a single Bayesian recursion. However, in practice, existing numerical approximations cause reappearing objects to be initialized as new tracks, especially after long periods of being undetected. In occlusion handling, the filter’s efficacy is dictated by trade-offs between the sophistication of the occlusion model and computational demand. Our contribution is a novel modeling method that exploits object features to address reappearing objects whilst maintaining a linear complexity in the number of detections. Moreover, to improve the filter’s occlusion handling, we propose a fuzzy detection model that takes into consideration the overlapping areas between tracks and their sizes. We also develop a fast version of the filter to further reduce the computational time. The source code is publicly available at \href{https://github.com/linh-gist/VisualRFS}{https://github.com/linh-gist/VisualRFS}.
\end{abstract}

}

\maketitle

\IEEEdisplaynontitleabstractindextext

%
\IEEEpeerreviewmaketitle

\section{Introduction}
The aim of multi-object tracking (MOT) in computer vision is to estimate the trajectories of multiple objects from an image sequence. This long-standing problem has a host of applications including surveillance, anomaly detection, developmental biology, and robotics. The most popular methodology is tracking-by-detection \cite{bewley2016simple}, where the problem is decomposed into two major tasks: i) \textit{detection}, which identifies and locates objects in each video frame; and ii) \textit{association}, which matches objects and existing trajectories or initiates newly appeared trajectories. The main advantages of tracking-by-detection are the versatility with respect to various detectors, and  good trade-offs between computational efficiency and performance.

Video data are rich in information that could be exploited to improve tracking by reducing data association uncertainty and resolving object occlusion and appearance-reappearance. The traditional practice of using simple information (e.g., bounding boxes) is not sufficient for good association between frames \cite{bewley2016simple}, and hence additional information is needed to improve tracking performance. With the advent of efficient machine learning techniques, more and more visual tracking solutions are exploiting sophisticated information from video data than the simple bounding boxes.

In track initialization/termination, resolving track appearance-reappearance is a challenging problem. When an existing track is undetected for a certain duration, either due to occlusion or leaving the scene, it is usually terminated, and then incorrectly initialized as a new track if it reappears. Thus, resolving appearance-reappearance and occlusion are two interrelated problems. In most visual tracking techniques, occlusion also causes identity (ID) switching, especially in crowded scenes \cite{dendorfer2020mot20}, and persistent occlusions.

While multiple hypothesis tracking has traditionally been the most popular in visual MOT \cite{B006}, the random finite set (RFS) framework has gained considerable attention due to its versatility, efficiency, and direct conceptual parallels with standard Bayesian state estimation
\cite{vo2013labeled}. Using a finite marked point process model with distinct marks, commonly known as labeled random finite set (LRFS), MOT filters integrate the sub-tasks of track management, state estimation, clutter rejection, and occlusion/miss-detection handling into a single Bayesian recursion. Several RFS-based MOT filters have been successfully applied to visual tracking \cite{fu2018particle,kim2019labeled,abbaspour2021online}. In its exact form, the generalized labeled multi-Bernoulli (GLMB) filter \cite{vo2013labeled,vo2014labeled} optimally addresses object appearance, disappearance, reappearance, and occlusion in a Bayesian sense. However, existing numerical approximations, designed to reduce computations (and memory), resulted in the initialization of reappearing objects as new tracks, and hence increased ID switching. In addition, optimal occlusion handling is compromised in real-time applications by sacrificing the level of sophistication in the detection model for computational speed.

\begin{figure}[t!]
    \centering
    \includegraphics[width=0.5\textwidth]{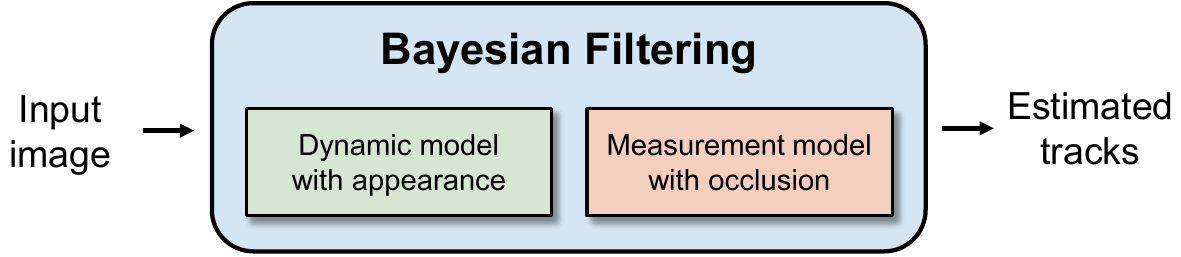}
    \caption{A high-level diagram of the proposed LRFS-based MOT algorithm.}
    \label{fig:overview}
\end{figure}

Using the LRFS framework, this paper proposes an online visual MOT algorithm that resolves appearance-reappearance and occlusion on top of the standard MOT functionalities. The overview of the proposed method is described in Figure~\ref{fig:overview}. Conceptually, our MOT algorithm uses a convolutional neural network (CNN) to detect object locations and extract deep re-ID features, which are fed into a GLMB filter to generate tracks. The novelty of our solution lies in a novel implementation of the GLMB filter and the exploitation of object features to minimize the initialization of reappearing objects as new tracks. In addition, we improve the proposed MOT filter's occlusion handling by developing a fuzzy detection model, which considers the overlaps between tracks and those varying areas over time, whilst maintaining a linear complexity in the number of detections. The prudent use of object features and positions in the GLMB filter also reduces the data association uncertainty, which in turn reduces the number of ID switches. To further reduce computation time, we also devise a labeled multi-Bernoulli (LMB) approximation \cite{reuter2014labeled} of the GLMB filter without significant sacrifice on tracking accuracy. Extensive performance evaluation of the proposed algorithms, on well-known tracking benchmarks, shows a lower number of ID switches and tracking errors compared to other state-of-the-art (SOTA) trackers.

In summary, our main contributions are as follows:
\begin{itemize}
  \item We design multi-object dynamic and measurement models under the LRFS framework. Our approach includes an appearance model using object features and occlusion handling based on fuzzy logic.
  \item Under the proposed dynamic and measurement models, we develop a visual multi-object tracker, based on GLMB filtering recursion, that can manage track initialization and re-ID. We also further develop a more efficient recursion using LMB approximation.
  \item We conduct extensive experiments to demonstrate the proposed methods on MOT benchmarks. Specifically, we evaluate the performance of our LRFS filtering solutions against the SOTA methods on MOT16, MOT17, and MOT20 datasets.
\end{itemize}

The rest of this paper is organized as follows. We introduce the related works in Section~\ref{sec:backgrounds}. Section~\ref{sec:models} presents the proposed models. In Section~\ref{sec:bayesian}, we present our Bayesian filtering recursions. Section~\ref{sec:implementation} describes implementation details, and Section~\ref{sec:experiment} evaluates the performance of the proposed method. Finally, Section~\ref{sec:conclusion} concludes this paper.

\section{Related Work}\label{sec:backgrounds}

\subsection{Visual Multi-Object Tracking}
\subsubsection{Batch/Online Multi-Object Tracking}
    Visual tracking can be performed in \textit{Batch} or \textit{Online}. Batch MOT produces tracks offline, i.e., after the entire batch of data has been received, such as hierarchical track association \cite{huang2008robust}, global trajectory optimization via dynamic programming \cite{berclaz2006robust}, and network flow \cite{shitrit2013multi}. Online MOT produces tracks after receiving each datum, and hence more suitable for online applications than batch methods. 

\subsubsection{Using Features in  Multi-Object Tracking}
    Features from video data provide compelling information for improving visual tracking performance. Although simple (handcrafted) features have been broadly used in computer vision, it is the use of deep features that has shown impressive performance improvements \cite{nanni2017handcrafted}. POI~\cite{yu2016poi} builds a cost matrix for data association on the combination of motion, shape, and appearance affinity based on CNN. DeepSORT \cite{wojke2017simple} also employs CNN trained on a large-scale person re-identification dataset, or ByteTrack \cite{zhang2022bytetrack} that also uses low confidence score detections. Moreover, MOTDT \cite{chen2018real} considers unreliable detection by combining detection and tracking results as candidates and selecting optimal candidates based on CNN. Subsequently, its derivative MOT such as JDE \cite{wang2020towards} and YOLOTracker \cite{chan2022online} that use Darknet backbones for feature extraction, CSTrack \cite{liang2020rethinking} that enhances the collaborative learning between detection and feature extraction tasks, FairMOT \cite{zhang2021fairmot} that balances between detection and re-ID feature quality, and GSDT \cite{wang2020joint} that employs graph neutral network, have improved the tracking performance using deep features. Recently, SiamMT \cite{vaquero2022tracking} and SiamMOT trackers \cite{vaquero2023real} can track objects in real-time by omitting the object detection task and improving the efficiency of the feature extractor.
    
\subsubsection{Occlusion Handling}
    Occlusion is a challenging problem in MOT and can be formulated as a detection problem where detectors could be trained to detect different parts (segments) of an object \cite{Li2018multiple}. Alternatively, in \cite{koporec2023human}, occluded objects are detected with high-level reasoning using a hierarchical compositional model. However, miss-detection is usually encountered in severe or full occlusion. There are MOT algorithms that have separate modules designed specifically to handle occlusion. One solution is to model the object depth \cite{Ma2010DepthAO} to identify occlusion. Further, the integration of occlusion attention modules into the tracking schemes based on the spatiotemporal/spatial context among objects \cite{ong2020bayesian} or their interactions \cite{Ur-Rehman2016multi} is a popular trend in the literature.

\subsection{Visual RFS-based Localization and Tracking}
    Multi-object localization filters only estimate the states of the objects, and unlike the single-object case, a sequence of sets (of state estimates) does not provide a set of trajectory estimates. The most popular RFS localization method is the Probability Hypothesis Density (PHD) filter~\cite{vo2006gaussian}. In \cite{zhou2013game}, game theory was applied to resolve occlusion handling within the PHD filter. More recently, a particle PHD filter~\cite{fu2018particle} was proposed with enhanced adaptive gating and group-based dictionary learning. 

    MOT filters estimate the trajectories of the objects, which include their state estimates. Analogous to single-object tracking LRFS MOT filters are formulated from a single Bayesian recursion that allows multiple trajectory estimates to be constructed from a sequence of sets of state estimates. The GLMB \cite{vo2013labeled,vo2014labeled} filter and its one-term approximation the LMB filter \cite{reuter2014labeled} are representative LRFS MOT filters. In visual tracking, an online visual GLMB filter was proposed to combine detection and image observations in  \cite{kim2019labeled}. This is extended in \cite{ong2020bayesian} to multi-view 3-D tracking with a realistic occlusion model that accommodates lines of sight, mostly neglected in other tracking methods due to computational load. In \cite{abbaspour2021online}, new modules for handling occlusions and ID switches are applied to the GLMB filter using location and simple features.

\section{Dynamic and Measurement Models\label{sec:models}}

In this section, we propose the dynamic and measurement models that facilitate object appearance for track re-ID and occlusion handling. A list of important notations is given in Table~\ref{tab:notations}.
\begin{table}[h!]
\footnotesize
\renewcommand{\arraystretch}{1}
    \caption{List of important notations.\label{tab:notations}}
    \begin{centering}
    \begin{tabular}{cl}
    \toprule 
    {\footnotesize{}\textbf{Symbol}} & {\footnotesize{}\textbf{Description}}\tabularnewline
    \midrule
    {\footnotesize{}$p^{X}$} & {\footnotesize{}$\prod_{x\in X}p(x)$}\tabularnewline
    {\footnotesize{}$\delta_{Y}[X]$} & {\footnotesize{}General Dirac delta: 1 if $X=Y$, 0 otherwise}\tabularnewline
    {\footnotesize{}$1_{X}(x)$} & {\footnotesize{}Inclusion function: 1 if $x\in X$, 0 otherwise}\tabularnewline
    {\footnotesize{}$a^{T}$} & {\footnotesize{}Transpose of $a$}\tabularnewline
    {\footnotesize{}$I_{n}$} & {\footnotesize{}$n$-D identity matrix}\tabularnewline
    {\footnotesize{}$\otimes$} & {\footnotesize{}Kronecker product}\tabularnewline
    {\footnotesize{}diag$(\cdot)$} & {\footnotesize{}Convert vector to diagonal matrix}\tabularnewline
    {\footnotesize{}$\mathcal{N}(\cdot,m,P)$} & {\footnotesize{}Gaussian with mean $m$, covariance $P$}\tabularnewline
    {\footnotesize{}Subscript `+'} & {\footnotesize{}Denote next time step quantity}\tabularnewline
    \bottomrule
    \end{tabular}
    \par\end{centering}
\end{table}

\subsection{Multi-Object Dynamic and Appearance Model\label{subsec:dynamics-model}}

We follow the standard LRFS model, in which an object is represented by $\boldsymbol{x}=(x,\ell)$, where $x$ $\in \mathbb{X}$ is its unlabeled state, and $\ell$ $\in \mathbb{L}$ is its unique label. $\mathbb{X}$ and $\mathbb{L}$ are the state space and (discrete) label space, respectively.
Conventionally, a label has the form $\ell=[k,i]$, where $k$
is the time when the object is born and $i$ is a unique index to
distinguish it from objects born at the same time \cite{vo2013labeled}. 

Different from a standard tracking model, in which unlabeled state usually encapsulates only the kinematic information of the track, in this work, the unlabeled state also contains the track appearance feature. Specifically, an unlabeled state is given as $x^{(\alpha)}$ $= (\zeta, \sigma)$ where $\alpha$ is an appearance feature parameter (not a part of the object state), $\zeta$ is a kinematic component, and $\sigma$ is a discrete mode such that: $\sigma=0$ if there is no change in appearance; or $\sigma=1$ if the object changes its appearance. The kinematic and appearance of an object are assumed to be statistically independent. For brevity, we only include the superscript $\alpha$ when it is necessary.

The \emph{multi-object state} at the current time step is a set  $\boldsymbol{X}=\{\boldsymbol{x}_{1},...,\boldsymbol{x}_{n}\}$ of objects, and is modeled as a finite marked point process with distinct marks. This means that it is also simple, and hence, commonly known as LRFS \cite{vo2013labeled}. The multi-object state at the next time step is formed by thinning and superposition, as follows. Each object $\boldsymbol{x} \in \boldsymbol{X}$ can either survive with probability
$P_{S}(\boldsymbol{x})$  and evolves to state $\boldsymbol{x}_{+}$ 
with transition density $\boldsymbol{f}_{S,+}(\boldsymbol{x}_{+}|\boldsymbol{x})=f_{S,+}^{(\ell)}(x_{+}|x)\delta_{\ell}[\ell_{+}]$,
or it might disappear with probability $1-P_{S}(\boldsymbol{x})$.
Note that the label of an object is unchanged over the course of its existence. Further, a set $\boldsymbol{X}_{B}$ of new
objects (births) might also appear in the scene. For an object
$\boldsymbol{x}_{B}=(x_{B},\ell_{B})\in\boldsymbol{X}_{B}$, its
birth probability is $P_{B}^{(\ell_{B})}$, and its state has probability density $f_{B}^{(\ell_{B})}(x_{B})$. Effectively, a multi-object density of new births at the next time step can be written as $\{P_{B,+}^{(\ell)},f_{B,+}^{(\ell)}\}_{\ell\in\mathbb{B}_{+}}$.

Denote $\mathbb{B}_{k}$ the set of all birth labels at
time step $k$, the label space up to time $k$ is
$\mathbb{L}=\biguplus_{t=0}^{k}\mathbb{B}_{t}$. Hence, the multi-object
state $\boldsymbol{X}$ is a finite subset of the labeled state
space $\mathbb{X}\times\mathbb{L}$. Let $\mathcal{L}(x,\ell)\triangleq\ell$
and $\mathcal{L}(\boldsymbol{X})\triangleq\{\mathcal{L}(\boldsymbol{x}):\boldsymbol{x}\in\boldsymbol{X}\}$,
the labels of the multi-object state $\boldsymbol{X}$ is distinct if $\Delta(\boldsymbol{X})=1$,
where $\Delta(\boldsymbol{X})=\delta_{|\boldsymbol{X}|}[|\mathcal{L}(\boldsymbol{X})|]$.

\subsection{Measurement Model with Occlusion\label{subsec:detection-model}}
 Measurements are modeled by the thinning of false negatives and the superposition of false positives. We propose to capture the spatial relationship among the objects to explicitly model how objects occlude each other. This model is encapsulated in the multi-object measurement likelihood.
For a multi-object state $\boldsymbol{X}$, each 
$\boldsymbol{x}\in\boldsymbol{X}$ is either detected
with a probability $P_{D}(\boldsymbol{x}, \boldsymbol{X} \backslash \{\boldsymbol{x}\})$
\cite{ong2020bayesian} and generates a measurement $z$ (in a measurement space $\mathbb{Z}$), or miss-detected
with a probability $1-P_{D}(\boldsymbol{x}, \boldsymbol{X} \backslash \{\boldsymbol{x}\})$.
Different from the standard multi-object detection model
\cite{vo2016efficient}, to account for occlusion, the detection probability of an object $\boldsymbol{x}$
in this work depends on other objects in the multi-object state, i.e., $\boldsymbol{X} \backslash \{\boldsymbol{x}\}$. 

The current measurement set $Z\!\!=\!\!\{z_{1},..,z_{M}\}$ includes measurements generated by the objects and false
positives. The number of false
positives is modeled by a Poisson 
with mean $\lambda_{c}$, and the false
positives are assumed to be uniformly
distributed on $\mathbb{Z}.$ This measurement model is 
given by
\begin{equation}
    \boldsymbol{g}(Z|\boldsymbol{X})\propto\sum_{\theta\in\Theta}1_{\Theta(\mathcal{L}(\boldsymbol{X}))}(\theta)\left[\psi_{Z,\boldsymbol{X}}^{(\theta)}\right]^{\boldsymbol{X}},\label{eq:detection_model}
\end{equation}
where 
\begin{equation}
    \psi_{Z,\boldsymbol{X}}^{(\theta)}(\boldsymbol{x})\!=\!
    \begin{cases}
        \frac{P_{D}(\boldsymbol{x},\boldsymbol{X} \backslash \{\boldsymbol{x}\})g(z_{j}|\boldsymbol{x})}{e^{-\lambda_{c}}(V_{\mathbb{Z}})^{-1}},     &   j=\theta(\mathcal{L}(\boldsymbol{x}))>0\\
        1-P_{D}(\boldsymbol{x},\boldsymbol{X} \backslash \{\boldsymbol{x}\}),   &   \theta(\mathcal{L}(\boldsymbol{x}))=0
    \end{cases},
\end{equation}
$V_{\mathbb{Z}}$ is the volume of the measurement space $\mathbb{Z}$,
$g(z_{j}|\boldsymbol{x})$ is the single-object likelihood function,
and $\theta\in\Theta$ is an association map which maps the object
labels to: the index of the measurements in the measurement set $Z$ if the object is detected; 0 if the object is miss-detected.

Each measurement consists of a kinematic component $\gamma$ (a bounding box)
and an appearance feature $\varrho$, i.e., $z=(\gamma, \varrho)$. Since object appearance is assumed statistically independent from its kinematic, given an object $\boldsymbol{x}^{(\alpha)}=(\zeta, \sigma, \ell)$,
the single-object likelihood can be written in a separable form, i.e.,
\begin{equation}
    g(\gamma,\varrho|\zeta, \alpha, \sigma, \ell)=g(\gamma|\zeta, \ell)g(\varrho|\sigma,\alpha).
\end{equation}

If an object is detected, we update its feature such that $\alpha_{+}=0.9\alpha+0.1\varrho$,
where $\alpha$ is the current feature and $\alpha_+$ is the updated feature. Hence, the object feature at the current time is the moving average of features from the associated measurements from the time the object was initialized up to the current time step. This method of updating the feature is also used in \cite{wang2020towards}, which enhances the robustness of the feature. It prevents the object feature from changing drastically when the object is occluded (i.e., the observed feature is the occluder feature, not the object feature). If an object is miss-detected, its feature is unchanged.

\subsection{Fuzzy Detection Model}\label{subsec:fuzzy}
\begin{figure}[h!]
    \begin{centering}
    \includegraphics[width=0.5\textwidth]{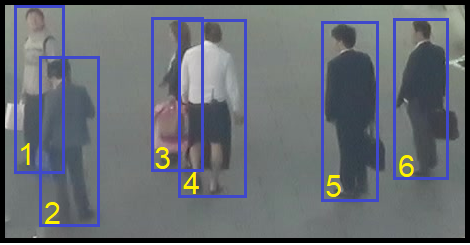}
    \par\end{centering}
    \caption{Tracks 1 and 2 overlap each other, but since the bottom corner of track 2 is lower than of track 1, track 2 occludes track 1. Similarly, track 4 occludes track 3.\label{fig:ioa-demo}}
\end{figure}

Object occlusion is modeled via the detection probability $P_{D}$ which depends on the amount of overlaps between
the object of interest and other objects in the scene. In single-view tracking, assuming that the camera is always
above the ground and objects move on the same ground level, the 
lower the bottom corner of the bounding boxes, the closer the corresponding objects are to the
camera, hence having higher chances of occluding neighboring objects (see Figure~\ref{fig:ioa-demo}). Given an object with state $x$, the amount of overlap with another object with state $x'$ can be measured by the intersection
over area (IoA) score \cite{abbaspour2021online}, 
\begin{equation}
    IoA(x,x')=\frac{\textrm{IntersectionArea}(x,x')}{\textrm{Area}(x)}. 
\end{equation}

Further, small objects (those far away from the camera) are usually difficult to detect. This can be integrated into the measurement model by introducing a size-dependent factor to the detection probability. Hence, the detection probability depends on both the ratio between the area of an object's bounding box and the average area of all objects. Given a set of objects $\boldsymbol{X}=\{\boldsymbol{x}_{1},...,\boldsymbol{x}_{n}\}$, the area ratio $R_a$ for an object $\boldsymbol{x}\in\boldsymbol{X}$ is defined as 
\begin{equation}\label{eq:Ra}
    R_{a}=\min\left(2, \frac{n\times\textrm{Area}(\boldsymbol{x})}{\sum_{j=1}^{n}\textrm{Area}(\boldsymbol{x}_{j})}\right),
\end{equation}
where Area$(\cdot)$ computes the area of an object from its state. The maximum IoA score for each object in the set can be computed according to Algorithm~\ref{alg:maxIoA}. 
\begin{algorithm}[h!]
    \small
    \caption{Computation of maximum IoA}
    \label{alg:maxIoA}
    
    \SetKwProg{Pn}{Function}{:}{\KwRet}
    \SetKwInOut{Input}{Input}
    \SetKwInOut{Output}{Output}
    \SetKwInOut{Procedure}{Procedure}

    \Input{$\boldsymbol{X}$}
    \Output{\textit{AllMaxIoA}}
    
    \vspace{0.1cm}\hrule\vspace{0.1cm}
    
    $\bar{\boldsymbol{X}} \leftarrow \boldsymbol{X}$\;
    \textit{AllMaxIoA} $\leftarrow 1 \times |\bar{\boldsymbol{X}}|$ zero list (accessed by $\ell$)\;
    \While{$|\bar{\boldsymbol{X}}| > 0$}{
        $\boldsymbol{x} \leftarrow$ an object (in $\bar{\boldsymbol{X}}$) having the lowest bottom corner\;
        $\bar{\boldsymbol{X}} \leftarrow \boldsymbol{\bar{X}\backslash \{x\} }$\;
        \For{$\boldsymbol{\bar{x}} \in \boldsymbol{\bar{X}}$}{
            \If{$IoA(x, \bar{x}) > AllMaxIoA[\ell]$}{
                $AllMaxIoA[\ell] \leftarrow IoA(x,\bar{x})$\;
            }
        }
    }
\end{algorithm}

\begin{figure}[h!]
    \begin{centering}
    \includegraphics[width=0.5\textwidth]{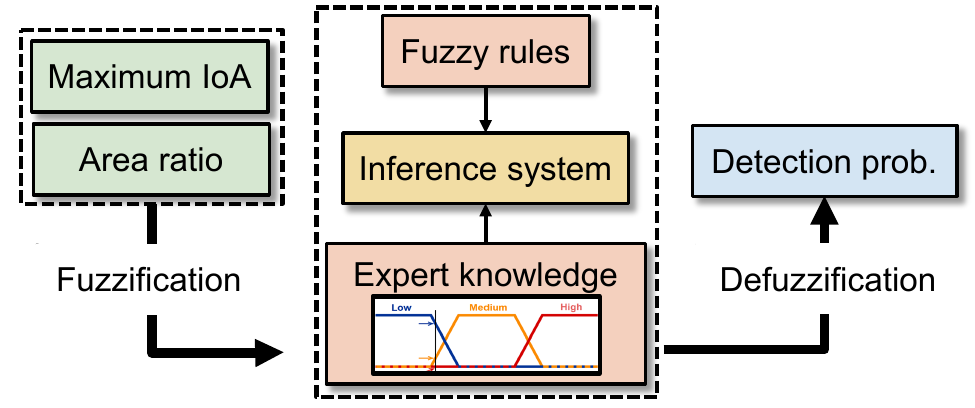}
    \par\end{centering}
    \caption{Design of our fuzzy detection model capable of handling object occlusion. The core of the model is a set of fuzzy rules and membership functions that represent expert knowledge. The inputs to the model are the maximum IoA score (computed using Algorithm \ref{alg:maxIoA}) and the area ratio (computed using (\ref{eq:Ra})). The output of the model is the object detection probability.\label{fig:fuzzy-system}}
\end{figure}

We propose a fuzzy detection model to establish the relationship between the degree of overlap, the object size, and the detection probability. Figure~\ref{fig:fuzzy-system} depicts the design of our fuzzy model with the input variables are the maximum IoA and the area ratio $R_{\text{a}}$, and the output variable is the object
detection probability.
\begin{figure}[h!]
    \begin{centering}
    \includegraphics[width=0.47\textwidth]{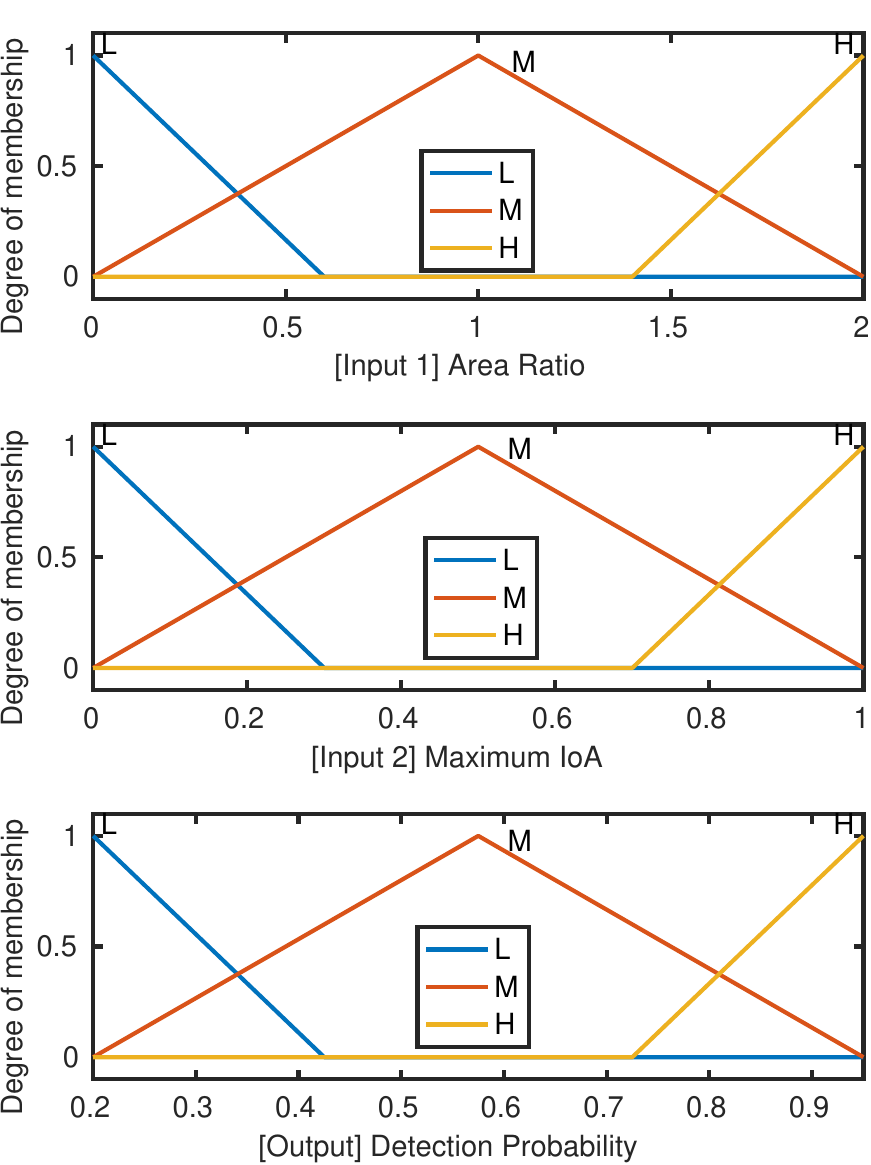}
    \par\end{centering}
    \caption{Membership functions for different degrees of membership (L, M, H). The range is limited to  {[}0, 2{]} for $R_{\text{a}}$, to {[}0, 1{]} for IoA, and
    {[}0.2, 0.99{]} for detection probability.\label{fig:fuzzy-membership-function}}
\end{figure}

Our fuzzy detection model has a collection of fuzzy sets with membership functions that represent expert knowledge. We design three fuzzy sets: low (L); medium (M); and high (H). Each sub-figure in Figure~\ref{fig:fuzzy-membership-function} shows the membership functions of the three fuzzy sets for each variable. For example, the first sub-figure shows the membership functions of the area ratio. For each area ratio (true) value (x-axis, ranging from 0 to 2 in our design), the plots in the sub-figure determine the degree of membership of the area ratio belonging to the fuzzy sets L, M or H (y-axis). For illustration, if the area ratio is 0.8, it has a good chance of belonging to the M set while having no chance of belonging to the L and H sets. Similar interpretations are for the sub-figures of the maximum IoA (value restricted between 0 and 1) and detection probability (value restricted between 0.2 and 0.95) variables.

\begin{table}[h!]
    \small
    \renewcommand{\arraystretch}{1}
    \caption{Fuzzy rules for detection probability.\label{tab:fuzzy-Relationship-between-membership}}
    \centering
    \begin{tabular}{|c||*{3}{c|}}
        \hline 
        \diagbox[height=0.8cm]{\textbf{Area ratio}}{\textbf{IoA}} 
        &\makebox[1em]{L}&\makebox[1em]{M}&\makebox[1em]{H}\\\hline\hline 
        L & M & L & L \\\hline 
        M & M & M & L \\\hline 
        H & H & H & L \\\hline
    \end{tabular}
\end{table}

\begin{figure}[h!]
    \begin{centering}
    \includegraphics[width=0.55\textwidth]{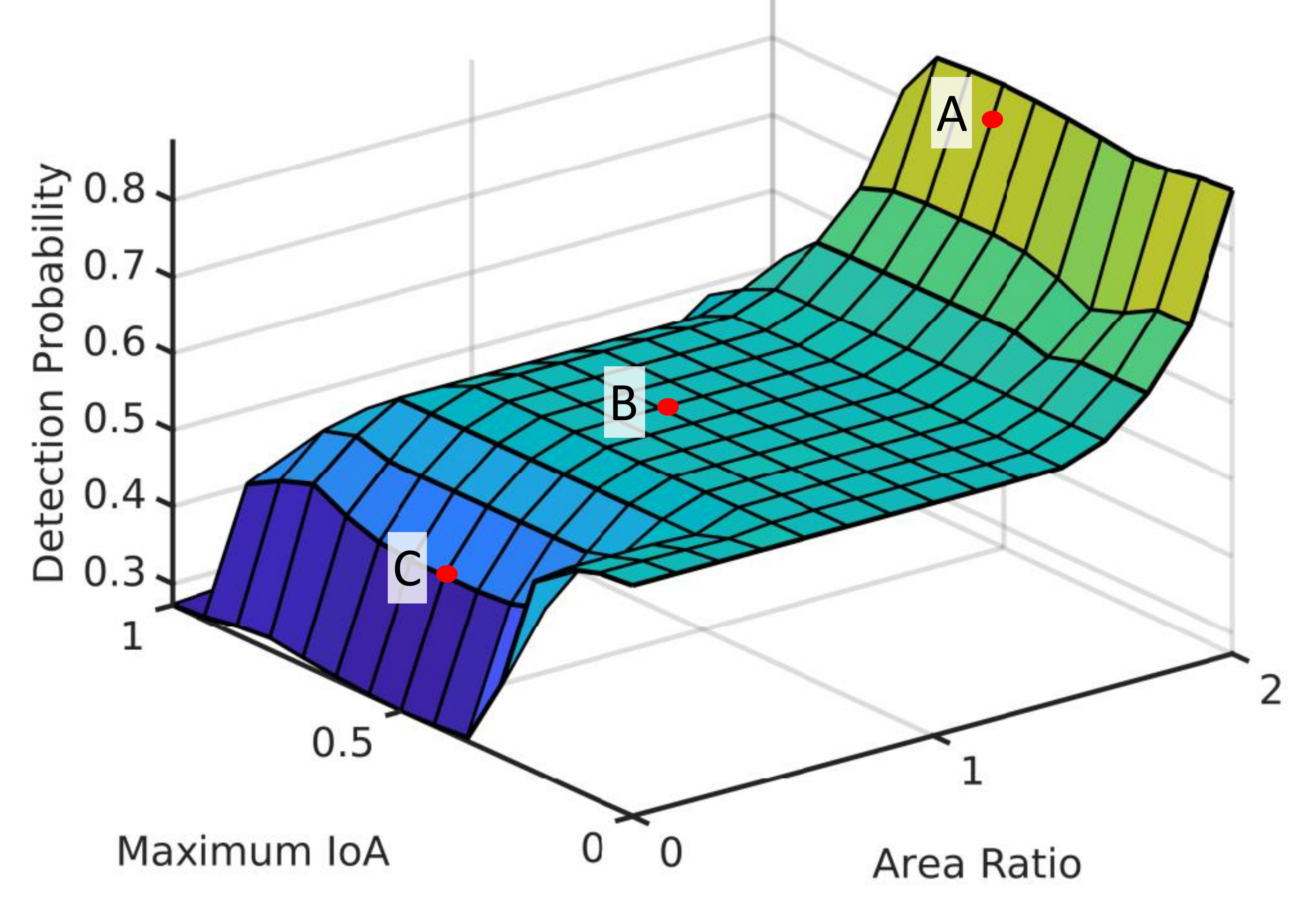}
    \par\end{centering}
    \caption{Relationship between variables in the fuzzy model. For illustration, we select 3 points on the plot that represent 3 typical occlusion scenarios: point A is the scenario where the object is covered half (IoA $=0.5$), but it is close to the camera (high area ratio), hence the high detection probability ($>0.9$); point B is the scenario where the object is also covered half, but it is relatively further away from the camera, hence the medium detection probability ($\approx0.5$); and point C is for the scenario where the object is small and far away from the camera, hence the low detection probability ($<0.5$).\label{fig:fis-surface}}
\end{figure}

Our fuzzy rule is designed based on the intuition that occluded objects and small objects have low detection probability. The fuzzy rule defined in Table~\ref{tab:fuzzy-Relationship-between-membership} reflects these intuitions where the trend of IoA score contradicts the trend of the detection probability, and the trend of $R_a$ follows that of the detection probability. For instance, if the IoA score is low, the detection probability is high; or if the $R_a$ score is low the detection probability is low.  Experimentally we observe that if the fuzzy rule follows a similar line to the presented intuition, the tracking performance is similar. The performance only decreases when the rule is counter-intuitive (see the ablation study in Subsection \ref{subsec:fuzzyabl}). The relationship between the detection probability and the amount of overlap and size given our fuzzy model is visualized in Figure~\ref{fig:fis-surface}. 

\section{Bayesian Multi-Object Filtering Solutions}\label{sec:bayesian}
\begin{figure*}[h!]
    \begin{centering}
    \includegraphics[width=\textwidth]{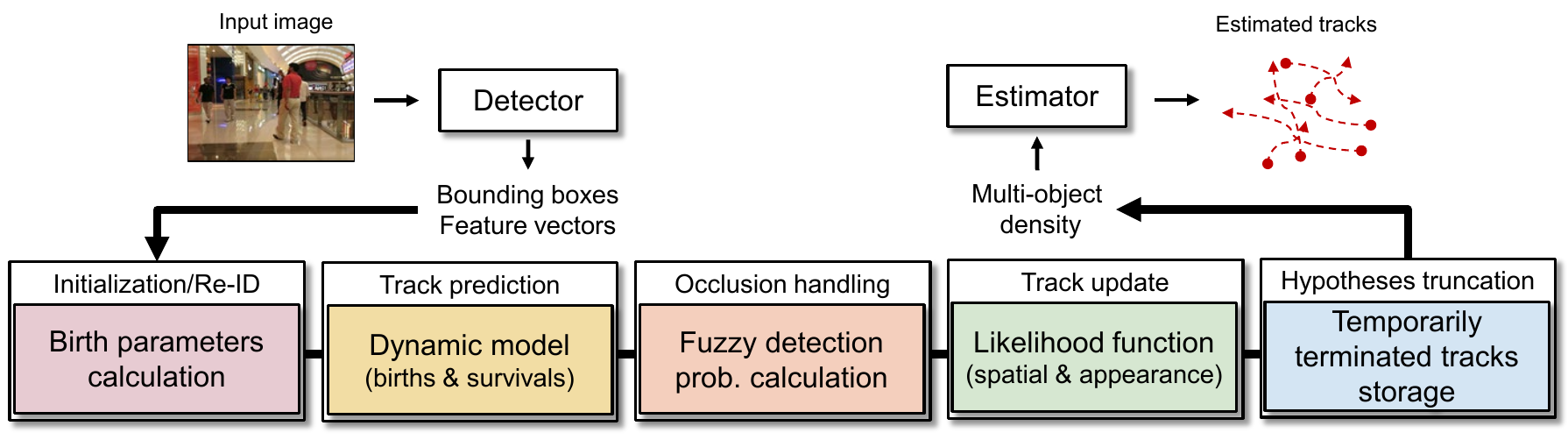}
    \par\end{centering}
    \caption{The proposed LRFS trackers exploiting object features to address track initialization/re-ID, and the fuzzy detection model to handle occlusions.
    \label{fig:tracker-diagram}}
\end{figure*}

In this section, we present an exact filtering recursion that is a direct result from applying Bayesian filtering formulation to our dynamic and measurement models. Nevertheless, since the implementation of this exact filtering recursion is intractable, we also propose practical approximations based on the GLMB and LMB filters. Figure~\ref{fig:tracker-diagram} illustrates the structure of our trackers.

\subsection{The Exact Filtering Recursion\label{subsec:exact}}
A general labeled multi-object density which encapsulates all information of a multi-object state can be written as
\begin{equation}
    \boldsymbol{\pi}\left(\boldsymbol{X}\right)=\Delta\left(\boldsymbol{X}\right)\sum_{I,\xi}\omega^{\left(I,\xi\right)}\delta_{I}[\mathcal{L}\left(\boldsymbol{X}\right)]p^{(\xi)}(\boldsymbol{X}),\label{eq:mixture_prior}
\end{equation}
where each $I\subseteq\mathbb{L}$ is a set of labels, each $\xi$
is a history of association maps, $\omega^{\left(I,\xi\right)}$ is a non-negative
weight such that $\sum_{(I,\xi)}\omega^{\left(I,\xi\right)}=1$, and $p^{(\xi)}(\boldsymbol{X})$ is a function that satisfies
\begin{equation}
    \int p^{(\xi)}(\{({x}_{1},\ell_{1}),...,({x}_{n},\ell_{n})\})dx_{1:n}=1.
\end{equation}
Note that, $p^{(\xi)}(\boldsymbol{X})$ is the joint probability density function of the unlabeled state of $\boldsymbol{X}$ and it also captures the interaction among objects (e.g., due to occlusion). Hereon, we refer to $p^{(\xi)}(\boldsymbol{X})$ as the \emph{unlabeled multi-object state} density.

Under the Bayesian framework, given a prior $\boldsymbol{\pi}$, a multi-object dynamic transition $\boldsymbol{f}_{+}$, a multi-object measurement likelihood $\boldsymbol{g}_{+}$, and a set of new measurements $Z_{+}$,  the filtering density can be written as:
\begin{equation}
    \boldsymbol{\pi}_{+}\left(\boldsymbol{X}_{+}|Z_{+}\right) \hspace{0.1cm} \propto \hspace{0.1cm} \boldsymbol{g}_{+}\left(Z_{+}|\boldsymbol{X}_{+}\right)\textstyle{\int}\boldsymbol{f}_{+}\!\left(\boldsymbol{X}_{+}|\boldsymbol{X}\right)\boldsymbol{\pi}\left(\boldsymbol{X}\right)\delta\boldsymbol{X}.\label{e:MTBF-1}    
\end{equation}

With the prior of the form given in (\ref{eq:mixture_prior}) and the multi-object dynamic and measurement models proposed in Section \ref{sec:models}, a direct application of the above Bayesian recursion yields the following filtering density, 
\begin{multline}
\Omega^{(E)}\left(\boldsymbol{\pi},\{P_{B,+}^{(\ell)},f_{B,+}^{(\ell)}\}_{\ell\in\mathbb{B}_{+}},Z_{+}\right)\propto \Delta(\boldsymbol{X}_{+})\\
\sum_{I,\xi,I_{+},\theta_{+}}\!\!\!\delta_{I_{+}}[\mathcal{L}(\boldsymbol{X}_{+})]\omega_{Z_{+},\boldsymbol{X}_{+}}^{(I,\xi,I_{+},\theta_{+})}p_{Z_{+}}^{(\xi,\theta_{+})}(\boldsymbol{X}_{+}).
\end{multline}
Details of this recursion are given in Section 1.1 of supplementary materials.

Nevertheless, the implementation of this filtering recursion may be impractical since storing and propagating the unlabeled multi-object state density are expensive given the number of hypothesis components ${(I,\xi)}$ grows exponentially over time. An alternative filtering solution is to propagate a GLMB or an LMB density that well approximates the general labeled multi-object density. It then allows the application of efficient filtering recursions \cite{vo2016efficient,reuter2014labeled}. In the next subsections, we establish two approximations, i.e., GLMB and LMB filters, which are suitable for real-time visual tracking applications.

\subsection{The GLMB Filter\label{subsec:glmb}}
Let us assume the initial multi-object states can be written in the form of a GLMB density \cite{vo2013labeled}, i.e.,
\begin{equation}
    \boldsymbol{\pi}\left(\boldsymbol{X}\right)=\Delta\left(\boldsymbol{X}\right)\sum_{I,\xi}\omega^{\left(I,\xi\right)}\delta_{I}[\mathcal{L}\left(\boldsymbol{X}\right)]\left[p^{(\xi)}\right]^{\boldsymbol{X}},\label{eq:GLMB_prior}
\end{equation}
where $p^{(\xi)}(\cdot,\ell)$
is a probability density on $\mathbb{X}$. Its compact form can be written as $\{ \left(\omega^{\left(I,\xi\right)},p^{\left(\xi\right)}\right):\left(I,\xi\right)\in\allowbreak\mathcal{F}(\mathbb{L})\times\Xi\}$. Different from (\ref{eq:mixture_prior}), the GLMB density does not capture the interaction among objects since each object is assumed statistically independent from each other, i.e., the unlabeled multi-object state density can be written in a separable form.

Even though the probability density for each object state in the GLMB prior is statistically independent from each other, the dependence among elements of the set $\boldsymbol{X}_{+}$ is induced due to the detection probability term $P_{D}(\boldsymbol{x}_{+}, \boldsymbol{X}_{+}\backslash\{\boldsymbol{x}_{+}\})$ in the measurement model. To facilitate the approximation, we assume that the states of objects in $\boldsymbol{X}_{+}$ concentrate around the estimates (e.g., means or modes), the detection probability of an object can be written as a constant $P_{D}(\hat{\boldsymbol{x}}_{+};\hat{\boldsymbol{X}}_{+})$, where $\hat{\boldsymbol{X}}_{+}\!\!=\!\!\{(\hat{x}_{+},\ell)\!:\!\ell\in\mathcal{L}(\boldsymbol{X}_{+})\backslash\mathcal{L}(\boldsymbol{x}_{+})\}$, and $\hat{x}_{+}$ is the estimate of the prediction density $p_{+}^{(\xi)}(x_{+},\ell)$. Hence, the multi-object filtering density can be written in terms of a GLMB density
\begin{multline}
\Omega^{(G)}\left(\boldsymbol{\pi},\{P_{B,+}^{(\ell)},f_{B,+}^{(\ell)}\}_{\ell\in\mathbb{B}_{+}},Z_{+}\right)\propto \Delta(\boldsymbol{X}_{+}) \\
\!\!\!\sum_{I,\xi,I_{+},\theta_{+}}\!\delta_{I_{+}}[\mathcal{L}(\boldsymbol{X}_{+})]\omega_{Z_{+},\hat{\boldsymbol{X}}_{+}^{(\xi,I_{+})}}^{(I,\xi,I_{+},\theta_{+})}[p_{Z_{+},\hat{\boldsymbol{X}}_{+}^{(\xi,I_{+})}}^{(\xi,\theta_{+})}(\cdot)]^{\boldsymbol{X}_{+}},\label{eq:GLMB_filtering}
\end{multline}
where $\hat{\boldsymbol{X}}_{+}^{(\xi,I_{+})}=\{(\hat{x}_{+}^{(\xi,\ell)},\ell):\ell\in I_{+}\}$, and $\hat{x}_{+}^{(\xi,\ell)}$ is the estimate from the prediction density $p_{+}^{(\xi)}(\cdot,\ell)$. Details of this filtering recursion are given in Section 1.2 of supplementary materials.

Efficient implementation of the GLMB filtering recursion allows direct selection of
the set of significant $(I_{+},\theta_{+})$ from a prior hypothesis
$(I,\xi)$ (referred to as joint prediction-update strategy \cite{vo2016efficient})
via Gibbs sampling or solving a rank assignment problem. However, observe the filtering density (\ref{eq:GLMB_filtering}),
$\hat{\boldsymbol{X}}_{+}^{(\xi,I_{+})}$ is not available to us until the set $I_{+}$ is selected. Hence, we cannot construct a sampling distribution or cost matrix to select significant $(I_{+},\theta_{+})$ if we do not have $I_{+}$ beforehand.

To circumvent this issue, note that $\hat{\boldsymbol{X}}_{+}^{(\xi,I_{+})}$ appears due to the detection probability term in the multi-object detection model. Following \cite{ong2020bayesian}, we use the set $\hat{\boldsymbol{X}}_{+}^{(\xi,I\cup\mathbb{B}_{+})}$ to compute the \emph{pseudo detection probability} of an object $\boldsymbol{x}=(x,\ell)$, i.e.,  $P_{D}(\hat{\boldsymbol{x}};\hat{\boldsymbol{X}}_{+}^{(\xi,L)})$, where $L=(I\cup\mathbb{B}_{+})\backslash\{\ell\}$. It allows us to construct an approximate cost matrix to select significant children hypotheses. Given the selected $(I_{+},\theta_{+})$, we then compute the correct hypothesis
weights. Further, the area ratio $R_{a}$ (input of the fuzzy model that computes the detection probability) can be approximated by the ratio between the estimated area from prediction density (of the object of interest) and the average area of estimated objects from the previous time step.

\subsection{The LMB Filter\label{subsec:lmb}}
We can further increase the efficiency of the above GLMB recursion by approximating the GLMB density by an LMB density, i.e., a one-term GLMB of the form  
\begin{equation}
    \boldsymbol{\pi}(\boldsymbol{X})  =  \Delta(\boldsymbol{X})\prod_{i\in\mathbb{L}}(1-r^{(i)})\prod_{\ell\in\mathcal{L}(\boldsymbol{X})}\frac{1_{\mathbb{L}}(\ell)r^{(\ell)}}{1-r^{(\ell)}}[p]^{\boldsymbol{X}},\label{eq:LMB_prior}
\end{equation}
where $r^{(\ell)}$ is the existence probability of an object with label
$\ell$,
and $p(\cdot,\ell)$ is the probability density of its state. For compactness, an LMB density is written in terms
of its parameters as $\boldsymbol{\pi}\triangleq\left\{ \left(r^{(\ell)},p^{(\ell)}\right)\right\} _{\ell\in\mathbb{L}}$. 

Applying the same strategy as for the GLMB approximation, if the prior density is an LMB density, we have the filtering density of the form
\begin{multline}
\Omega^{(L)}\left(\boldsymbol{\pi},\{P_{B,+}^{(\ell)},f_{B,+}^{(\ell)}\}_{\ell\in\mathbb{B}_{+}},Z_{+}\right) \propto \Delta(\boldsymbol{X}_{+}) \\
\sum_{I_{+},\theta_{+}}\delta_{I_{+}}[\mathcal{L}(\boldsymbol{X}_{+})]\omega_{Z_{+},\hat{\boldsymbol{X}}_{+}^{(I_{+})}}^{(I_{+},\theta_{+})}[p_{Z_{+},\hat{\boldsymbol{X}}_{+}^{(I_{+})}}^{(\theta_{+})}(\cdot)]^{\boldsymbol{X}_{+}},\label{eq:exact_GLMB_LMB_filtering}
\end{multline}
where $\hat{\boldsymbol{X}}_{+}^{(I_{+})}=\{(\hat{x}_{+}^{(\ell)},\ell):\ell\in I_{+}\}$, and $\hat{x}_{+}^{(\ell)}$ is the estimate from the
prediction density $p_{+}(\cdot,\ell)$. The details of this filtering recursion are given in Section 1.3 of supplementary materials.

Note that the filtering density in (\ref{eq:exact_GLMB_LMB_filtering}) takes on the form of a GLMB. For closed-form recursion, we approximate this GLMB density with an LMB density that matches the 1st-moment and cardinality distribution. This LMB density can be obtained by aggregating the GLMB components to a single component \cite{reuter2014labeled}. Hence, the filtering density in (\ref{eq:exact_GLMB_LMB_filtering}) is then approximated
by an LMB density with an aggregator $\Lambda$, i.e.,
\begin{equation}
    \Lambda\!\left(\Omega\left(\boldsymbol{\pi},\{P_{B,+}^{(\ell)},f_{B,+}^{(\ell)}\}_{\ell\in\mathbb{B}_{+}},Z_{+}\right)\right)\!=\!\{(r^{(\ell)},p^{(\ell)})\}_{\ell\in\mathbb{L}_{+}},
\end{equation}
where
\begin{eqnarray}
r^{(\ell)} & \!\!\!\!\!=\!\!\!\!\! & \sum_{I_{+},\theta_{+}}\omega_{Z}^{(I_{+},\theta_{+})}1_{I_{+}}(\ell),\\
p^{(\ell)} & \!\!\!\!\!=\!\!\!\!\! & \frac{1}{r^{(\ell)}}\sum_{I_{+},\theta_{+}}\omega_{Z}^{(I_{+},\theta_{+})}1_{I_{+}}(\ell)p^{(\theta_{+})}(x,\ell).\label{eq:GLMB_to_LMB}
\end{eqnarray}

\emph{Remark.} The aggregator performs the summation over all hypotheses $(I_{+},\theta_{+})$ for each object labeled $\ell$. For a small number of hypotheses, the GLMB filter is more efficient since aggregation is more computationally expensive than actually propagating all hypotheses. Nevertheless, when the number of hypotheses is large, the efficiency of the LMB filter is exhibited.

\subsection{Track Initialization and Re-ID}\label{subsec:reID}
In visual tracking, a track is assigned a new ID only when it first appears in the scene. Otherwise, reappeared tracks should regain their original ID (re-ID). The exact filter or its GLMB and LMB approximations can implicitly handle track initialization and (theoretically) track reappearance without any explicit modules. However, due to the truncation process for tractable computation (where only significant association hypotheses are selected), reappeared tracks after a long period of miss-detection are usually assigned different ID (since the original tracks are discarded when their probability of existence is not significant). 

To handle track reappearance while keeping the filter tractable, when tracks are terminated by the filter (due to truncation), we store them in a separate
memory in order to recall them if they later reappear in the scene (referred to as temporarily terminated (TT) tracks). In particular, we store their IDs and appearance features. If a track belongs to different hypotheses at the time step before it is removed from the GLMB density, we store its feature in the most significant hypothesis, among the ones that contain it, at that time step. We only use the appearance feature for re-ID. TT tracks are only disregarded completely when they are not recalled after a long period of time (e.g., 50 frames).

Our track initialization/re-ID is based on the adaptive birth model in \cite{reuter2014labeled}, i.e., current time step measurements are used to initialize tracks at the next time step. The intuition of this model is when a measurement is not associated with known tracks, it is likely to be generated by a new birth (or reappeared track). Hence, this model initializes or recalls tracks with one time step delay. 

At the end of the filtering cycle at time step $k$, we compute the association probability of the $j^{th}$ measurement $z_{j}$ as \cite{reuter2014labeled} 
\begin{equation}
    r_{U}(z_{j})=\sum_{I,\xi}1_{\xi^{(k)}}(j)\omega_{Z}^{\left(I,\xi\right)},\label{eq:meas-ass-prob}
\end{equation} 
where $\xi^{(k)}$ is the measurement index to track association map at time $k$. The existence probability of the (new/reappeared) track formed by this measurement at the next time step is computed as:
\begin{equation}
    \!\!P_{B,+}\left(\theta_{B,+}(j)\right) = \min\left(\!P_{B,max},\frac{1-r_{U}(z_{j})}{\sum_{\gamma\in Z}1-r_{U}(\gamma)}\lambda_{B}\!\right)\!,
\end{equation}
where $P_{B,max}<1$ is a constant, $\lambda_{B}$ is the number of possible new/reappeared tracks per time step, and $\theta_{B,+}$ is a bijective map that maps each measurement index to an $\ell\in\mathbb{B}_{+}$. Note that for the LMB filter, the measurement association probabilities (\ref{eq:meas-ass-prob}) are computed from the GLMB filtering density before applying GLMB to LMB conversion (aggregator) $\Lambda$.

Prior to computing the filtering density, we start the track initialization by checking for track reappearance. From the measurement set at the current time step, we only consider measurements with low association probability (i.e., less than a threshold $\tau_{B}$), we denote this set of measurements as $Z_{B}$ (i.e., $Z_{B} = \{z|r_{U}(z)<\tau_{B}$, $z\in Z$\}). For each measurement $z_{B}\in Z_{B}$ (associated with a feature vector $\varrho_{B}$) and each TT track (associated with a feature vector $\alpha$), we compute the cosine similarity between $\varrho_{B}$ and $\alpha$. The track is recalled only when the cosine similarity is greater than some threshold. The recall process is performed such that tracks with high cosine similarity scores are recalled first. The remaining measurements in $Z_{B}$ are used to initialize tracks with new ID when all TT tracks are recalled, or the cosine similarity condition cannot be met anymore.

$P_{B,+}$ is used as the existence probability of the new/reappeared track. The state probability density of a track $f_{B,+}$ and its appearance feature are obtained from $z_{B}$. That gives us the multi-object density of the new births  $\{P_{B,+}^{(\ell)},f_{B,+}^{(\ell)}\}_{\ell\in\mathbb{B}_{+}}$ discussed in Section \ref{subsec:dynamics-model}.

\subsection{Multi-Object Estimator\label{subsec:Estimator}}
Given the filtering density $\boldsymbol{\pi}_{+}$,
the multi-object estimators are used to extract the estimated tracks. For GLMB density parameterized by $\{ \left(\omega^{\left(I,\xi\right)},p^{\left(\xi\right)}\right):\left(I,\xi\right)\in\allowbreak\mathcal{F}(\mathbb{L})\times\Xi\}$, we first compute the cardinality distribution
(the probability distribution on the number of objects) via $\rho(n)=\sum_{I,\xi}\omega^{(I,\xi)}\delta_{n}[|I|]$ in \cite{vo2013labeled}, then the estimated cardinality is $\hat{N}=\textrm{argmax}(\rho(n)).$
The estimated hypothesis is $(\hat{I},\hat{\xi})=\arg\max_{(I,\xi)}(\omega^{(I,\xi)}\delta_{\hat{N}}[|I|])$,
and the set of estimated objects is $\hat{\boldsymbol{X}}=\{(\hat{x},\ell):\ell\in\hat{I}\}$ with $\hat{x}=\arg\max_{x}(p^{(\hat{\xi})}(x,\ell))$ in \cite{vo2013labeled}.

For an LMB density parameterized by $\{r^{(\ell)},p^{(\ell)}\}_{\ell\in\mathbb{L}},$
the set of estimated objects is $\hat{\boldsymbol{X}}=\{(\hat{x},\ell):r_{max}^{(\ell)}>\theta_{u}\textrm{ and }r^{(\ell)}>\theta_{l}\}$ with $\hat{x}=\arg\max(p^{(\ell)})$ \cite{reuter2014labeled}. $\theta_{u}$ and $\theta_{l}$ are the upper and lower existence
probability thresholds, respectively, and $r_{max}^{(\ell)}$ is the
maximum existence probability of an object labeled $\ell$.

\section{Implementation Details}\label{sec:implementation}
In this section, we provide details on the implementation of the proposed GLMB and LMB trackers.
\subsection{Object Dynamic Representation}\label{subsec:imp-dynamic-models}
We model the kinematic state of an object with a bounding box moving in 2-D image, i.e., a state is presented by an 8-D vector $\zeta=[u,\dot{u},v,\dot{v},h,\dot{h},\beta,\dot{\beta}]^{T}$,
where $u$ and $v$ are the horizontal and vertical coordinates of the
box centroid, $h$ is the height of the box, $\beta$ is the aspect ratio of the box (height over
width), $\dot{u}, \dot{v}, \dot{\beta}, \dot{h}$ are their respective rates of change. The constant velocity model is used for the object dynamic, i.e., 
$f_{S,+}^{(\ell)}(\zeta_{+}|\zeta)=\mathcal{N}\left(\zeta_{+};F\zeta,Q\right)$, where $
F=I_{4}\otimes\left[\begin{array}{cc}
1 & T\\
0 & 1
\end{array}\right],Q=\textrm{diag}(\epsilon_{Q})\otimes\left[\begin{array}{c}
\frac{T^{2}}{2}\\
T
\end{array}\right]\left[\begin{array}{cc}
\frac{T^{2}}{2} & T\end{array}\right] 
$, $T$ is the sampling period, and $\epsilon_{Q}$ is the noise variance vector. 
At prediction, we assume $p_{+}(\sigma_{+}=0)=0.9$ and $p_{+}(\sigma_{+}=1)=0.1$,
which means it is likely that the object's appearance does not change at consecutive time steps.

The longer an object is alive, the lower probability it disappears. At time $k$, we set the relationship between object's lifespan and its surviving probability such that \cite{kim2019labeled}

\begin{equation}
    P_{S}(x,\ell) = P_{S}(\zeta,\ell) = \frac{b(\zeta)}{\left(1 + \exp(-\tau_{L}(k - \ell[1,0]^{T}))\right)}, 
\end{equation}
where $b(\zeta)$ is a mask to set $P_{S}$ to high value where objects are likely to exist, or to low value where objects are likely to disappear, $k-\ell[1,0]^{T}$ is the temporal length of the object, and $\tau_{L}$ is a constant scaling factor.  Further, $b(\zeta)$ can also be used to lower surviving probability of small size tracks with negative rate of change (in size) since those tracks are indeed moving out of the scene. Specifically, $b(\zeta)$ can be written as

\begin{equation}
    b(\zeta) =
    \begin{cases}
        \widehat{P}_{S}/\left(1+\exp\left(-\tau_{S}\left(k_{S}-\frac{\widehat{\beta}\times\zeta_{5}{}^{2}}{\bar{A}}\right)\right)\right) & \!\!\textrm{if }\zeta_{6}<0\\
        \widehat{P}_{S} & \!\!\textrm{otherwise}
    \end{cases},
\end{equation}
where: $\zeta_{i}$ is the $i^{th}$ component of vector $\zeta$;  $\widehat{\beta}=\max(\beta_{\textrm{min}},\zeta_{7})$ with $\beta_{\textrm{min}}>0$ to ensure positive value area; $\widehat{P}_{S} \in (0,1]$ is constant surviving probability; $\tau_{S}$ and $k_{S}$ are constants controlling the scaling factor; $\bar{A}$ is the average area of the estimated objects in the previous time step. We assume the surviving probability of objects is uniform over the entire image. Hence, a constant $\widehat{P}_{S}$ is used rather than a function of object position. Further, assuming the probability density of $\zeta$ concentrates around its estimated value $\hat{\zeta}$, the surviving probability of a track is approximated by a constant $\bar{P}_{S}(\ell)$. 
\subsection{Single-Object Detection Model}\label{subsec:imp-detection-model}
The likelihood of observing the kinematic component $\gamma$ of a measurement $z$ has a Gaussian distribution form: $g(\gamma|\zeta,\ell)\!=\!\mathcal{N}\left(\gamma;H\zeta,R\right)$,
where $H\!=\!I_{4}\otimes\left[1 \thinspace\thinspace\thinspace 0\right],R\!=\!\textrm{diag}(\epsilon_{R})$, and $\epsilon_{R}$ is the observation noise variance vector $\epsilon_{R}$. Moreover, the likelihood of the appearance mode is defined as
\begin{eqnarray}
    g(\varrho|\sigma\!=\!0,\alpha) & \hspace{-0.1cm} = \hspace{-0.1cm} & d_{c}(\varrho,\alpha)^{\phi},\\ g(\varrho|\sigma\!=\!1,\alpha) & \hspace{-0.1cm} = \hspace{-0.1cm} & (1-d_{c}(\varrho,\alpha))^{\phi},
\end{eqnarray}where $d_{c}$ is the Euclidean distance between two vectors. Experimentally,
we observe $\phi=15$ best suits our visual tracking task. A measurement can be assigned to an 
object if its Mahalanobis distance (to the kinematic distribution), or the cosine dissimilarity between its feature and the object appearance feature is lower than prescribed thresholds.

\subsection{Hypothesis Truncation for GLMB/LMB Filters}\label{subsec:glmb-implementation}
Hypothesis truncation can be performed in a brute-force manner by generating all possible hypotheses and deleting the insignificant ones. Nevertheless, calculating the weights of all hypotheses is computationally prohibitive. Alternatively, the truncation process can be cast as an M-best assignment problem, which tractably selects hypotheses with significant weights, by setting the cost matrix to reflect the hypothesis weight. The cost matrix can be set up as follows. Denote the pseudo detection probability as $\hat{P}_{D}(\ell)$, if
a track parameterized with a feature vector $\alpha$ is associated with a measurement $z$ having a feature
vector $\varrho$, its data-updated probability density and weight are given respectively as 
\begin{eqnarray*}
p_{z}^{(\xi)}(\zeta_{+},\ell) & \!\!\!\!\!=\!\!\!\!\! & \mathcal{N}(\zeta,\bar{\zeta}_{+}+K(z-H\bar{\zeta}_{+}),[I_{8}-KH]P_{+}),\\
\bar{\psi}_{z}^{(\xi)}(\ell) & \!\!\!\!\!=\!\!\!\!\! & \hat{P}_{D}(\ell)q(\gamma)(p_{+}^{(\xi)}(\sigma_{+}=0)d_{cos}(\varrho,\alpha)^{\phi} \\
&\!\!\!\!\!& +p_{+}^{(\xi)}(\sigma_{+}=1)(1-d_{cos}(\varrho,\alpha))^{\phi}),
\end{eqnarray*}
where $q(\gamma)=\mathcal{N}(\gamma,H\bar{\zeta}_{+},HPH^{T}+R)$ and $K\!=\!PH^{T}[HPH+R]^{-1}$.
For a miss-detected track, its data-updated probability density
is the same as the prediction density, and its weight
is $\bar{\psi}_{0}^{(\xi,\theta_{+})}(\ell)=1\!-\!\hat{P}_{D}(\ell)$.

For each prior hypothesis $(I,\xi)$, let us enumerate $I$ as $I=\{\ell_{1:N}\}$,
the set of reappeared and new tracks as $\mathbb{N}_{+}=\{\ell_{N+1:P}\}$, and the measurement set as $Z=\{z_{1:M}\}$, the element at row $i^{th}$, column $j^{th}$ of a $P$ rows by $(M+2P)$ columns cost matrix is given as
\begin{equation}
    C_{i,j}=\begin{cases}
        -\ln \eta_{i}(j) & \textrm{if }j\in\{1:M\}\\
        -\ln \eta_{i}(0) & \textrm{if }j=M+i\\
        -\ln \eta_{i}(-1) & \textrm{if }j=M+P+i\\
    \infty & \textrm{otherwise}
    \end{cases},
\end{equation}
where
\begin{equation}
    \eta_{i}(j)=\begin{cases}
    1-\bar{P}_{S}^{(\xi)}(\ell_{i}) & \textrm{if }1\leq i\leq N,j<0\\
    \bar{P}_{S}^{(\xi)}(\ell_{i})\bar{\psi}_{z_{j}}^{(\xi)}(\ell_{i}) & \textrm{if }1\leq i\leq N,j>0\\
    \bar{P}_{S}^{(\xi)}(\ell_{i})\bar{\psi}_{0}^{(\xi)}(\ell_{i}) & \textrm{if }1\leq i\leq N,j=0\\
    1-P_{B}(\ell_{i}) & \textrm{if }N+1\leq i\leq P,j<0\\
    P_{B}(\ell_{i})\bar{\psi}_{z_{j}}^{(\xi)}(\ell_{i}) & \textrm{if }N+1\leq i\leq P,j>0\\
    P_{B}(\ell_{i})\bar{\psi}_{0}^{(\xi)}(\ell_{i}) & \textrm{if }N+1\leq i\leq P,j=0
    \end{cases}.
\end{equation}
Rank assignment algorithm \cite{vo2014labeled}  or Gibbs sampling \cite{vo2016efficient} is then applied to select assignment matrices that have low costs. Each selected assignment matrix (consisting of 0 or 1) must have every row summing to 1, and every column summing to either 0 or 1. These assignment matrices are then converted to hypotheses $(I_{+},\theta_{+})$, which are used to compute the hypothesis weights. Note that we only need to recompute the data-updated weight by replacing the pseudo detection probability $\hat{P}_{D}$ with the correct ${P}_{D}$ (compute using selected $I_{+}$), the data-updated probability densities of the object state remain the same (as the ones used to compute the cost matrix). The same implementation is applied for the LMB filter by considering the LMB density as a GLMB density with a single term.

\section{Experiments}\label{sec:experiment}
In this section, we evaluate the performance of our filters against SOTA methods on MOT15, MOT16, MOT17 \cite{milan2016mot16}, and MOT20 \cite{dendorfer2020mot20} (MOTChallenge) datasets. Further, we also provide an analysis of the efficiency of our approximations and ablation studies for different components of our filters. 

\subsection{Evaluation of Tracking Accuracy}\label{subsec:exp-accuracy}

\subsubsection{Performance Measures}
We use the CLEAR MOT measure \cite{bernardin2008evaluating}, including MOTA, MT, ML, FP, FN and IDS scores; the IDF1 score \cite{ristani2016performance}; the HOTA score \cite{luiten2021hota}; and the OSPA\!$^{\texttt{(2)}}$\! metric \cite{beard2020asolution,rezatofighi2020trustworthy} to evaluate the tracking performance. Note that OSPA\!$^{\texttt{(2)}}$\! metric measures the distance between two sets of tracks which is the tracking error (lower OSPA\!$^{\texttt{(2)}}$\! distance is better performance, details on the metric are given in Section 2 of supplementary materials).

\subsubsection{Parameter Setting}
The number of measurement association hypotheses is 500. Variance of dynamic noise is set to $\epsilon_{Q}=[9,9,9,10^{-4}]$ and the observation noise variance is set to $\epsilon_{R}=[50,50,50,10^{-3}]$. A measurement $z$ at the current time step is used to form new/reappeared tracks at the next time step if its association probability $r_{U}(z)<0.95$. The Poisson mean of the number of false measurements ($\lambda_c$) is chosen depending on the quality of the detector.

\subsubsection{Comparison with SOTA Methods on Validation Sets}
We compare results from our filters and ones from different SOTA methods, i.e., DeepSORT \cite{wojke2017simple}, JDE \cite{wang2020towards}, FairMOT \cite{zhang2021fairmot}, GSDT \cite{wang2020joint}, CSTrack~\cite{liang2020rethinking}, and TraDeS~\cite{wu2021track}.
For a fair comparison, we use the same detection results from the SOTA methods (obtained from pre-trained models provided by the authors) for our filters. Specifically, POI and FairMOT provide 128-D re-ID feature. JDE, CSTrack, and  GSDT provide 512-D re-ID feature. Generally, our methods exhibit better performance in terms of CLEAR MOT, IDF1, and HOTA scores, and significantly lower OSPA\!$^{\texttt{(2)}}$ errors as shown in Tables~\ref{tbl:differ_detectors_mot16}, \ref{tbl:differ_detectors_mot17} and \ref{tbl:mot20_val}. Overall, trackers using detection results from CSTrack and FairMOT achieve better performance due to a higher level of detection accuracy with strong discriminative features. The GLMB filter demonstrates better tracking performance compared to the LMB because the GLMB filter can handle multiple data association hypotheses in a more complete manner.

\begin{table*}[h!]
    \centering
    \renewcommand{\arraystretch}{1}
    \caption{Tracking results on MOT16 validation dataset (\textcolor{red}{\textbf{red}}: the best, \textcolor{blue}{\textbf{blue}}: the second best, \textbf{bold}: the best in a detector, `$\ast$': our methods, D.T.: default tracker of the detector).}
    \label{tbl:differ_detectors_mot16}
    \scriptsize
    \begin{tabular}{p{0.2cm}|p{1.9cm}| P{0.8cm} P{0.8cm} P{0.8cm} P{0.8cm} P{0.8cm} P{0.8cm} P{1cm}}
        \toprule
        \multirow{2}{*}{{\rotatebox{90}{\textbf{Det.}}}} & \multirow{2}{*}{\textbf{Method}} & \multirow{2}{*}{\textbf{MOTA$\uparrow$}} & \multirow{2}{*}{\textbf{IDF1$\uparrow$}} &  \multirow{2}{*}{\textbf{HOTA$\uparrow$}} & \multirow{2}{*}{\textbf{FP$\downarrow$}} & \multirow{2}{*}{\textbf{FN$\downarrow$}} & \multirow{2}{*}{\textbf{IDS$\downarrow$}} & \multirow{2}{*}{\textbf{OSPA\!$^{\texttt{(2)}}$\!$\downarrow$}} \\
        \multirow{1}{*}{} & & & \\   
        \midrule
            \multirow{3}{*}{{\rotatebox{90}{POI}}}
            & DeepSORT \cite{wojke2017simple}  &   \textbf{60.3} &   64.8   &   53.1 & 6,484 &   36,627 &   576 &   44.9\\
            & \textbf{LMB*} &  59.3 & \textbf{65.9} & \textbf{54.0} & 8,111 & \textbf{36,223} & \textbf{570} & 38.5\\
            & \textbf{GLMB*}    & 59.9 & 64.2 & 52.5 & \textbf{4,373} & 39,295 & 585 & \textbf{38.2} \\
        \midrule
            \multirow{3}{*}{{\rotatebox{90}{JDE}}}
            & D.T. \cite{wang2020towards} &   70.1 &   69.3   &   57.0 &   \textbf{5,929} &   25,927 &   1,160 &   48.2\\
            & \textbf{LMB*} &    70.5 & 66.7 & 56.0 & 12,606 & \textbf{19,025} & \textbf{960} & 36.2 \\
            & \textbf{GLMB*} &  \textbf{73.1} & \textbf{69.3} & \textbf{57.6} & 7,036 & 21,527 & 1,134 & \textbf{35.5}\\
        \midrule
            \multirow{3}{*}{{\rotatebox{90}{TraDes}} }
            & D.T. \cite{wu2021track} &   \textbf{71.1} &  \textbf{68.9}   &   \textbf{59.5} &   \textbf{4,263} &   26,972 &   638 &   48.0\\
            & \textbf{LMB*}  &   70.1 & 68.2 & 58.9 & 6,090 & \textbf{26,525} & \textcolor{blue}{\textbf{379}} & \textbf{33.9}\\
            & \textbf{GLMB*} &  70.1 & 66.7 & 57.8 & 4,306 & 28,235 & 495 & 34.0 \\ 
        \midrule
            \multirow{3}{*}{{\rotatebox{90}{CSTrack}}}
            & D.T. \cite{liang2020rethinking} &   79.0 &  79.6   &   68.4 &   \textcolor{blue}{\textbf{3,927}} &   18,635 &   581 &   36.0    \\
            & \textbf{LMB*}  &  77.8 & 82.7 & 72.4 & 13,541 & \textcolor{red}{\textbf{10,500}} & 465 & 27.9\\
            & \textbf{GLMB*}  &  \textcolor{blue}{\textbf{79.7}} & \textcolor{red}{\textbf{84.2}} & \textcolor{red}{\textbf{73.3}} & 6,786 & 14,969 & \textcolor{red}{\textbf{340}} & \textcolor{red}{\textbf{21.7}}\\ 
        \midrule
            \multirow{3}{*}{{\rotatebox{90}{FairMOT}}}
            & D.T. \cite{zhang2021fairmot} &   80.7 &  81.5  &   66.1 &   3,233 &   17,707 &  \textbf{409}  &   28.4\\
            & \textbf{LMB*}  &  \textcolor{red}{\textbf{81.9}} & 82.1 & \textcolor{blue}{\textbf{67.3}} & 5,663 & \textcolor{blue}{\textbf{13,844}} & 503 &  \textbf{26.3}\\
            & \textbf{GLMB*}  &  81.3 & \textcolor{blue}{\textbf{83.1}} & 67.2 & \textcolor{red}{\textbf{3,094}} & 17,121 & 439 & 27.3\\ 
        \midrule
            \multirow{3}{*}{{\rotatebox{90}{GSDT}}}
            & D.T. \cite{wang2020joint} &   70.8 &   71.2   &   59.1 &   10,098 &   \textbf{21,461} &   645 &   35.4\\
            & \textbf{LMB*}  &   72.0 & 75.4 & 61.8 & 8,454 & 22,007 & 463 & 26.4\\
            & \textbf{GLMB*}  &  \textbf{72.4} & \textbf{77.4} & \textbf{62.9} & \textbf{5,891} & 24,131 & \textbf{451} & \textcolor{blue}{\textbf{23.9}}\\
        \bottomrule
    \end{tabular}
    
\end{table*}

\begin{table*}[t!]
    \centering
    \renewcommand{\arraystretch}{1}
    \caption{Tracking results on MOT17 validation dataset (\textcolor{red}{\textbf{red}}: the best, \textcolor{blue}{\textbf{blue}}: the second best, \textbf{bold}: the best in a detector, `$\ast$': our methods, D.T.: default tracker of the detector).}
    \label{tbl:differ_detectors_mot17}
    \scriptsize
    \begin{tabular}{p{0.2cm}|p{1.9cm}|P{0.8cm} P{0.8cm} P{0.8cm} P{0.8cm} P{0.8cm} P{0.8cm} P{1cm}}
        \toprule
        \multirow{2}{*}{{\rotatebox{90}{\textbf{Det.}}}} & \multirow{2}{*}{\textbf{Method}} & \multirow{2}{*}{\textbf{MOTA$\uparrow$}} & \multirow{2}{*}{\textbf{IDF1$\uparrow$}} &  \multirow{2}{*}{\textbf{HOTA$\uparrow$}} & \multirow{2}{*}{\textbf{FP$\downarrow$}} & \multirow{2}{*}{\textbf{FN$\downarrow$}} & \multirow{2}{*}{\textbf{IDS$\downarrow$}} & \multirow{2}{*}{\textbf{OSPA\!$^{\texttt{(2)}}$\!$\downarrow$}} \\
        \multirow{1}{*}{} & & & \\ 
        \midrule
            \multirow{3}{*}{{\rotatebox{90}{POI}}}
            & DeepSORT \cite{wojke2017simple}  & \textbf{59.9} &   64.3   &   52.8 & 18,204 &   115,176  &   \textbf{1,740} &   43.1\\
            & \textbf{LMB*} &  59.0 & \textbf{65.6} & \textbf{53.7} & 23,232 & \textbf{113,259} & 1,761 & 36.3\\
            & \textbf{GLMB*} & 59.4 & 63.8 & 52.2 & \textbf{12,306} & 122,748 & 1,803 & \textbf{36.1}\\
        \midrule
            \multirow{3}{*}{{\rotatebox{90}{JDE}}}
            & D.T. \cite{wang2020towards} &  71.0 &  69.4   &   57.1 &   \textbf{14,202} &   79,860 &   3,597 &   45.0\\
            & \textbf{LMB*} &  71.9 & 67.1 & 56.4 & 33,351 & \textbf{58,272} & \textbf{2,949} & 34.0\\
            & \textbf{GLMB*} &  \textbf{74.2} & \textbf{69.5} & \textbf{57.9} & 17,070 & 66,207 & 3,531 & \textbf{33.6}\\
        \midrule
            \multirow{3}{*}{{\rotatebox{90}{TraDes}}}
            & D.T. \cite{wu2021track} &   \textbf{71.2} &  \textbf{68.6}   &   \textbf{59.3} & 10,578 &   84,387 &   1,974 &  44.9\\
            & \textbf{LMB*}  &  70.1 & 68.0 & 58.7 & 16,206 & \textbf{83,208} & \textcolor{blue}{\textbf{1,170}} & 34.6\\
            & \textbf{GLMB*} & 70.1 & 66.5 & 57.6 & \textbf{10,851} & 88,326 & 1,542 & \textbf{33.8} \\ 
        \midrule
            \multirow{3}{*}{{\rotatebox{90}{CSTrack}}}
            & D.T. \cite{liang2020rethinking} &  80.2 &   79.8   &   68.7 &   \textcolor{blue}{\textbf{7,512}} &   57,306 &   1,797 &   34.0\\
            & \textbf{LMB*}  &  79.5 & 83.4 & 73.0 & 35,439 & \textcolor{red}{\textbf{31,980}} & 1,485 & 25.9 \\
            & \textbf{GLMB*}  &  \textcolor{blue}{\textbf{81.2}} & \textcolor{red}{\textbf{85.0}} & \textcolor{red}{\textbf{74.0}} & 15,981 & 46,194 & \textcolor{red}{\textbf{1,053}} & \textcolor{red}{\textbf{20.6}} \\ 
        \midrule
            \multirow{3}{*}{{\rotatebox{90}{FairMOT}}}
            & D.T. \cite{zhang2021fairmot} &   81.1 &   81.5 & 66.0 &   6,678 &   55,785 &   \textbf{1,275} &   26.5\\
            & \textbf{LMB*}  &  \textcolor{red}{\textbf{82.3}} & 82.2 & \textcolor{blue}{\textbf{67.3}} & 13,911 & \textcolor{blue}{\textbf{44,124}} & 1,566 & \textbf{24.3} \\
            & \textbf{GLMB*}  & 81.9 & \textcolor{blue}{\textbf{82.9}} & \textcolor{blue}{\textbf{67.3}} & \textcolor{red}{\textbf{6,639}} & 52,833 & 1,512 & 25.5 \\ 
        \midrule
            \multirow{3}{*}{{\rotatebox{90}{GSDT}}}
            & D.T. \cite{wang2020joint} &  71.5 &   71.6   &   59.3 &   27,000 &   \textbf{66,801} & 2,145 &   30.1\\
            & \textbf{LMB*}  &  72.2 & 75.2 & 61.7 & 23,010 & 69,333 & 1,464 & 23.9\\
            & \textbf{GLMB*}  & \textbf{72.5} & \textbf{77.2} & \textbf{62.8} & \textbf{15,411} & 75,807 & \textbf{1,449} & \textcolor{blue}{\textbf{22.2}}\\
        \bottomrule
    \end{tabular}
    
\end{table*}

\begin{table*}[h!]
    \centering
    \renewcommand{\arraystretch}{1}
    \caption{Tracking results on MOT20 validation dataset (\textcolor{red}{\textbf{red}}: the best, \textbf{bold}: the best in a detector, `$\ast$': our methods, D.T.: default tracker of the detector).}
    \label{tbl:mot20_val}
    \scriptsize
    \begin{tabular}{p{0.2cm}|p{1.9cm}|P{0.8cm} P{0.8cm} P{0.8cm} P{0.8cm} P{0.8cm} P{0.8cm} P{1cm}}
        \toprule
        \multirow{2}{*}{{\rotatebox{90}{\textbf{Det.}}}} & \multirow{2}{*}{\textbf{Method}} & \multirow{2}{*}{\textbf{MOTA$\uparrow$}} & \multirow{2}{*}{\textbf{IDF1$\uparrow$}} &  \multirow{2}{*}{\textbf{HOTA$\uparrow$}} & \multirow{2}{*}{\textbf{FP$\downarrow$}} & \multirow{2}{*}{\textbf{FN$\downarrow$}} & \multirow{2}{*}{\textbf{IDS$\downarrow$}} & \multirow{2}{*}{\textbf{OSPA\!$^{\texttt{(2)}}$\!$\downarrow$}} \\
        \multirow{1}{*}{} & & & \\
        \midrule
            \multirow{3}{*}{{\rotatebox{90}{FairMOT}}}
             & D.T. \cite{zhang2021fairmot} &   75.5 &   79.1   &   70.5 & 51,837 &   271,682 &   4,540 &   56.2\\
            & \textbf{LMB*}  &   \textcolor{red}{\textbf{76.8}}  & 79.1  & 71.3 & \textbf{26,867}  & 281,669  & \textcolor{red}{\textbf{1,842}}  & \textcolor{red}{\textbf{33.3}}\\
            & \textbf{GLMB*}  &  \textcolor{red}{\textbf{76.8}}  & \textcolor{red}{\textbf{79.2}}  & \textcolor{red}{\textbf{71.5}} & 41,079  & \textcolor{red}{\textbf{266,721}}  & 2,508  & 42.4\\ 
            \midrule
            \multirow{3}{*}{{\rotatebox{90}{GSDT}}}
             & D.T. \cite{wang2020joint} & 74.5 &   76.3   &   67.7 &  \textcolor{red}{\textbf{21,974}} &   316,653 &   2,906 &   \textbf{37.2}\\
            & \textbf{LMB*}  &   75.5  & 77.1  & 69.0  & 28,733  & 296,926  & \textbf{2,299}  & 39.8\\
            & \textbf{GLMB*}  &  \textbf{76.0}  & \textbf{78.0}  & \textbf{70.2}  & 29,810  & \textbf{288,777}  & 2,440  & 40.1\\
        \bottomrule
    \end{tabular}
\end{table*}

\begin{table*}[t]
\centering
\renewcommand{\arraystretch}{1}
\caption{Result on MOTChallenge test sets (\textcolor{red}{\textbf{red}}: the best, \textcolor{blue}{\textbf{blue}}: the second best, `$\ast$': our methods).}
\label{tbl:testset}
    \scriptsize
    \begin{tabular}{p{1.9cm}| P{0.8cm} P{0.8cm} P{0.8cm} P{0.8cm} P{0.6cm} P{0.6cm} P{1cm} P{0.6cm}}
        \toprule
         \textbf{Method} & \textbf{MOTA$\uparrow$} & \textbf{IDF1$\uparrow$} &  \textbf{HOTA$\uparrow$} & \textbf{MT$\uparrow$} & \textbf{ML$\downarrow$} &  \textbf{FP$\downarrow$} & \textbf{FN$\downarrow$} & \textbf{IDS$\downarrow$}\\
        \midrule
           & \multicolumn{7}{c}{\textbf{MOT16 Test Set}}   \\
            
        \midrule
              POI \cite{yu2016poi} &   68.2 &   60.0   &  50.1 &  41.0 &  19.0  &  11,479 &  45,605 &   933 \\
              MOTDT \cite{chen2018real} &   47.6 &   50.9   &  - &   15.2 &   38.3  &  9,253 &  85,431 &   792 \\
              DeepSORT \cite{wojke2017simple} &   61.4 &   62.2   &  50.1 &    32.8 &   18.2   & 12,852 &   56,668 &   \textcolor{blue}{\textbf{781}} \\
              JDE \cite{wang2020towards} &  64.4 &    55.8   &   - &   35.4 &   20.0   &   - &   - &   1544 \\
              TraDes \cite{wu2021track} &   70.1 &   64.7   &  53.2 &   37.3 &   20.0   &   \textcolor{blue}{\textbf{8,091}} &   45,210 &   1,144\\
              CSTrack \cite{liang2020rethinking} &   \textcolor{red}{\textbf{75.6}} &   \textcolor{red}{\textbf{73.3}}  &  \textcolor{red}{\textbf{59.8}} &   42.8 &   16.5   &   9,646	 &   \textcolor{red}{\textbf{33,777}}	 &   1,121 \\
              FairMOT \cite{zhang2021fairmot} & 74.9 &   \textcolor{blue}{\textbf{72.8}}  &   - &   \textcolor{red}{\textbf{44.7}} &    \textcolor{blue}{\textbf{15.9}}  &   - &   - &   1,074\\
              GSDT \cite{wang2020joint} &   74.5 &   68.1  &  56.6 &   41.2 &   17.3  &   8,913 &   36,428	 &   1229 \\
              \textbf{LMB*}  &  73.1 & 71.7 & 59.0 & 42.4 & 21.1 & \textcolor{red}{\textbf{7,741}} & 40,562 & \textcolor{red}{\textbf{702}} \\ 
              \textbf{GLMB*}  &   \textcolor{blue}{\textbf{75.0}} & 72.4 & \textcolor{blue}{\textbf{59.4}} & \textcolor{blue}{\textbf{44.3}} & \textcolor{red}{\textbf{15.5}} & 9,526 & \textcolor{blue}{\textbf{35,027}} & 1,042 \\ 
        \midrule
            & \multicolumn{7}{c}{\textbf{MOT17 Test Set}}   \\
        \midrule
              TraDes \cite{wu2021track} &   69.1 &  63.9   &  52.7 &   36.4 &   21.5  &   \textcolor{red}{\textbf{20,892}} &   150,060 &   3,555\\
              CSTrack \cite{liang2020rethinking} &   \textcolor{red}{\textbf{74.9}} &  \textcolor{red}{\textbf{72.6}}   &  \textcolor{blue}{\textbf{59.3}} &   41.5 &   17.5   &   \textcolor{blue}{\textbf{23,847}} &   \textcolor{red}{\textbf{114,303}} &   3,567 \\
              FairMOT \cite{zhang2021fairmot} &   73.7 & \textcolor{blue}{\textbf{72.3}} &  \textcolor{red}{\textbf{59.3}} &  \textcolor{red}{\textbf{43.2}} &   \textcolor{blue}{\textbf{17.3}}   &   27,507	 &   \textcolor{blue}{\textbf{117,477}} &   3,303 \\
              GSDT \cite{wang2020joint} &   73.2 &   66.5  & 55.2 &   41.7 &   17.5   &   26,397 &   120,666 &   3,891 \\
              \textbf{LMB*} & 71.3 & 70.6 & 58.3 & 40.5 & 23.3 & 21,918 & 137,739 & \textcolor{red}{\textbf{2,151}} \\ 
              \textbf{GLMB*} & \textcolor{blue}{\textbf{73.9}} & 71.5 & 58.9 & \textcolor{blue}{\textbf{42.8}} & \textcolor{red}{\textbf{16.9}} & 25,116 & 118,989 & \textcolor{blue}{\textbf{3,255}} \\ 
        \midrule
            & \multicolumn{7}{c}{\textbf{MOT20 Test Set}}   \\
        \midrule
              FairMOT \cite{zhang2021fairmot} &   61.8 &  67.3 & \textcolor{red}{\textbf{54.6}} &   \textcolor{red}{\textbf{68.8}} &   \textcolor{red}{\textbf{7.60}}   & 103,440 & \textcolor{red}{\textbf{88,901}} &  5,243\\
              GSDT \cite{wang2020joint} &    67.1 &   \textcolor{blue}{\textbf{67.5}}   &  53.6 &    53.1 &   13.2   &   31,507 &   135,395 &   3,133\\
              CSTrack \cite{liang2020rethinking} &    \textcolor{blue}{\textbf{66.6}} &  \textcolor{red}{\textbf{68.6}}  & 54.0 &    50.4 &  15.5  &   \textcolor{red}{\textbf{25,404}} &   144,358 &   3,196\\
              \textbf{LMB*} &   65.4 & 67.1 & 54.1 & \textcolor{blue}{\textbf{60.5}} & \textcolor{blue}{\textbf{12.3}} & 57,919 & \textcolor{blue}{\textbf{118,337}} & \textcolor{red}{\textbf{2,787}} \\ 
              \textbf{GLMB*} &   \textcolor{red}{\textbf{67.7}} & 67.3 & \textcolor{blue}{\textbf{54.2}} & 54.7 & 14.5 & \textcolor{blue}{\textbf{29,597}} & 134,534 & \textcolor{blue}{\textbf{2,911}} \\ 
        \bottomrule
    \end{tabular}
    
\end{table*}
\subsubsection{Comparison with SOTA Methods on Test Sets}
In Table~\ref{tbl:testset}, we compare SOTA methods and ours on MOTChallenge test sets. We use GSDT detection results for MOT20, and FairMOT with 256-D for MOT16/17. All results are obtained from the MOTChallenge leaderboard. In this experiment, OSPA\!$^{\texttt{(2)}}$ error cannot be evaluated since we do not have the ground truth for test sets. In MOT16/17, ours are slightly worse than CSTrack (which is the best in this experiment), comparable to FairMOT, GSDT, and better than others. Our filters exhibit the lowest number of ID switches among other methods in this experiment, which indicates the utility of the proposed appearance-reappearance resolution and occlusion handling. 

Nevertheless, our methods have a slightly lower IDF1 score compared to the best value on each dataset. It is because the proposed trackers also include miss-detected tracks in their outputs. On the one hand, it helps maintain track continuity by not dropping the tracks in miss-detection events. It might also decrease the tracking accuracy if the estimate is far away from the ground truth (the estimated object state is not accurate if the prediction model has high uncertainty). In this case, this behavior is reflected by the low IDS (high track continuity), but also low IDF1 (low tracking accuracy).

Further, we observe that the LMB filter in MOT20 dataset has a high number of FP because of a high number of false tracks. This is due to the high number of false positive measurements and the drastic approximation of the LMB filter in this dataset (due to the large number of objects). In particular, the LMB approximation causes the object density to have high uncertainty. Since the uncertainty is high, false tracks could be assigned false positive measurements and have relatively high existence probability, hence being included in the set of output estimates.

\subsubsection{Comparison with SOTA LRFS Filters}

In Table~\ref{tbl:rfs_com}, we provide a comparison among LRFS filters including GLMB-IM \cite{kim2019labeled}, MOMOT \cite{abbaspour2021online}, and ours. The results from MOMOT tracker are taken from \cite{abbaspour2021online} (using public detection), whereas the results from ours and GLMB-IM filter are obtained by using FairMOT detector. Note that while our methods use object deep features from the FairMOT detector, GLMB-IM and MOMOT use handcrafted object features that are extracted from the image segment contained in the object bounding box.

It shows that our methods achieve significantly higher MOTA and IDF1 scores in MOT15 and MOT17 test sets. We note that our methods exhibit higher numbers of IDS compared to MOMOT. However, MOMOT has a lower number of correctly tracked objects (reflected by its low MT score), hence possibly the lower IDS number (unmatched tracks are counted as track loss or false positives, not as ID switches in the CLEAR MOT measure \cite{bernardin2008evaluating}).

\begin{table}[t!]
    \centering
    \caption{Results of SOTA LRFS filters and our methods on test sets (\textcolor{red}{\textbf{red}}: the best, `$\ast$': our methods).}
    \label{tbl:rfs_com}
    \scriptsize
    \begin{tabular}{l|l| c c c c c}
    \toprule
        \textbf{Dataset} & \textbf{Method} & \textbf{MOTA$\uparrow$} & \textbf{IDF1$\uparrow$}   & \textbf{MT$\uparrow$} & \textbf{ML$\downarrow$} & \textbf{IDS $\downarrow$}  \\ \midrule
        ~ & GLMB-IM \cite{kim2019labeled} & 29.1	 & 39.7 & 49.0 & \textcolor{red}{\textbf{9.8}} & 1,636   \\ 
        ~ & MOMOT \cite{abbaspour2021online} & 40.0 & 50.3 & 6.0 & 36.9 & \textcolor{red}{\textbf{307}}   \\ 
        MOT15 & \textbf{LMB*} & 55.3  & 62.3 & \textcolor{red}{\textbf{59.6}} & 10.0 & 614   \\ 
        ~ & \textbf{GLMB*} & \textcolor{red}{\textbf{59.9}} & \textcolor{red}{\textbf{62.4}} & 46.5 & 13.2 & 710  \\
        \midrule 
        ~ & GLMB-IM \cite{kim2019labeled} & 60.1 & 45.4		 & 21.5 & 32.7 & 6,381   \\ 
        ~ & MOMOT \cite{abbaspour2021online} & 55.5 & 63.4 & 19.0 & 35.9 & \textcolor{red}{\textbf{1,333}}   \\ 
        MOT17 & \textbf{LMB*} & 71.3 & 70.6 & 40.5 & 23.3 & 2,151   \\ 
        ~ & \textbf{GLMB*} & \textcolor{red}{\textbf{73.9}} & \textcolor{red}{\textbf{71.5}} & \textcolor{red}{\textbf{42.8}} & \textcolor{red}{\textbf{16.9}} & 3,255   \\
    \bottomrule
    \end{tabular}
\end{table}

\begin{figure*}[hp!]
    \centering
        \raisebox{0.5in}{\rotatebox[origin=lc]{90}{CSTrack}}
        \includegraphics[width=0.85\textwidth]{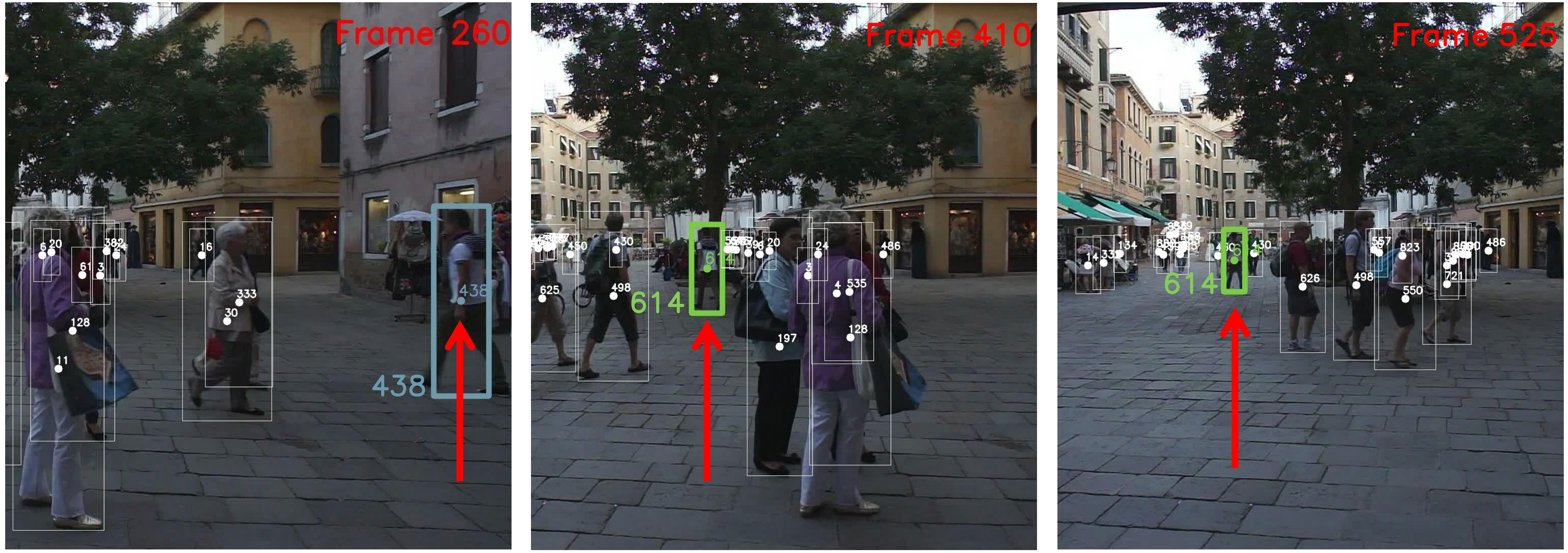}
        \raisebox{0.5in}{\rotatebox[origin=rc]{270}{MOT16-02}}
        
        \raisebox{0.5in}{\rotatebox[origin=lc]{90}{GLMB\_CSTrack}}
        \includegraphics[width=0.85\textwidth]{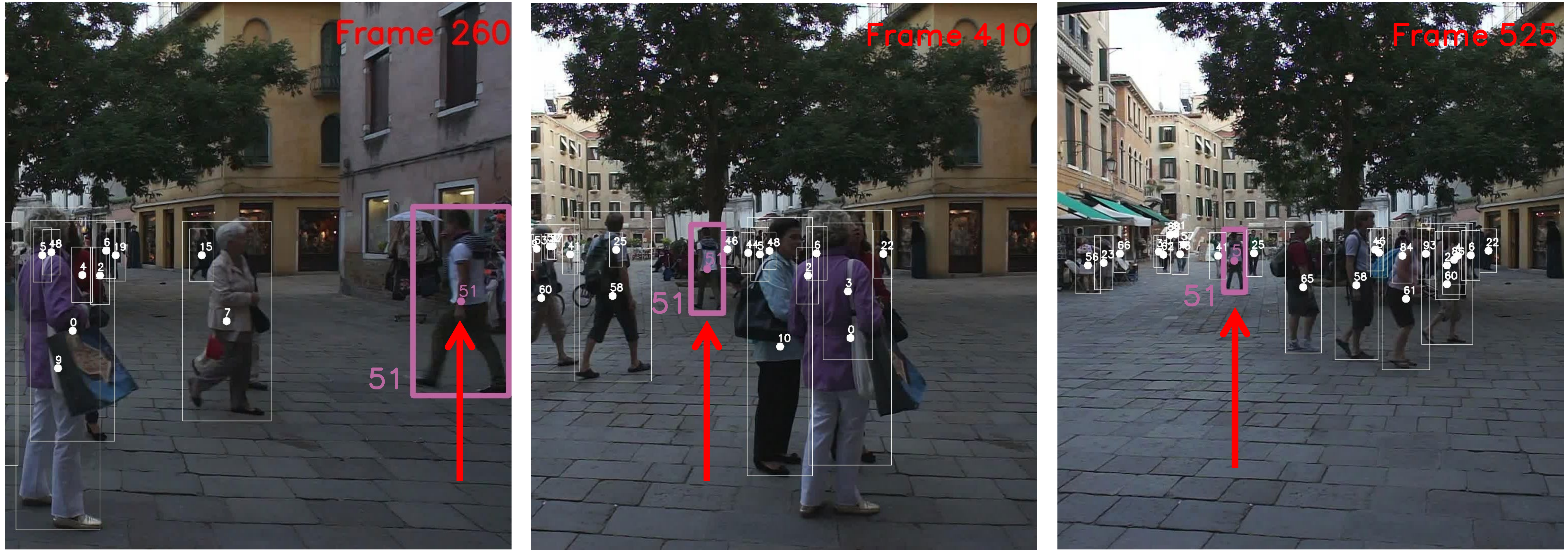}
        \raisebox{0.5in}{\rotatebox[origin=rc]{270}{MOT16-02}}\vspace{-0.53cm}
        
     \rule{0.925\textwidth}{.4pt}
    
        \vspace{0.06cm}
        \raisebox{0.5in}{\rotatebox[origin=lc]{90}{CSTrack}}
        \includegraphics[width=0.85\textwidth]{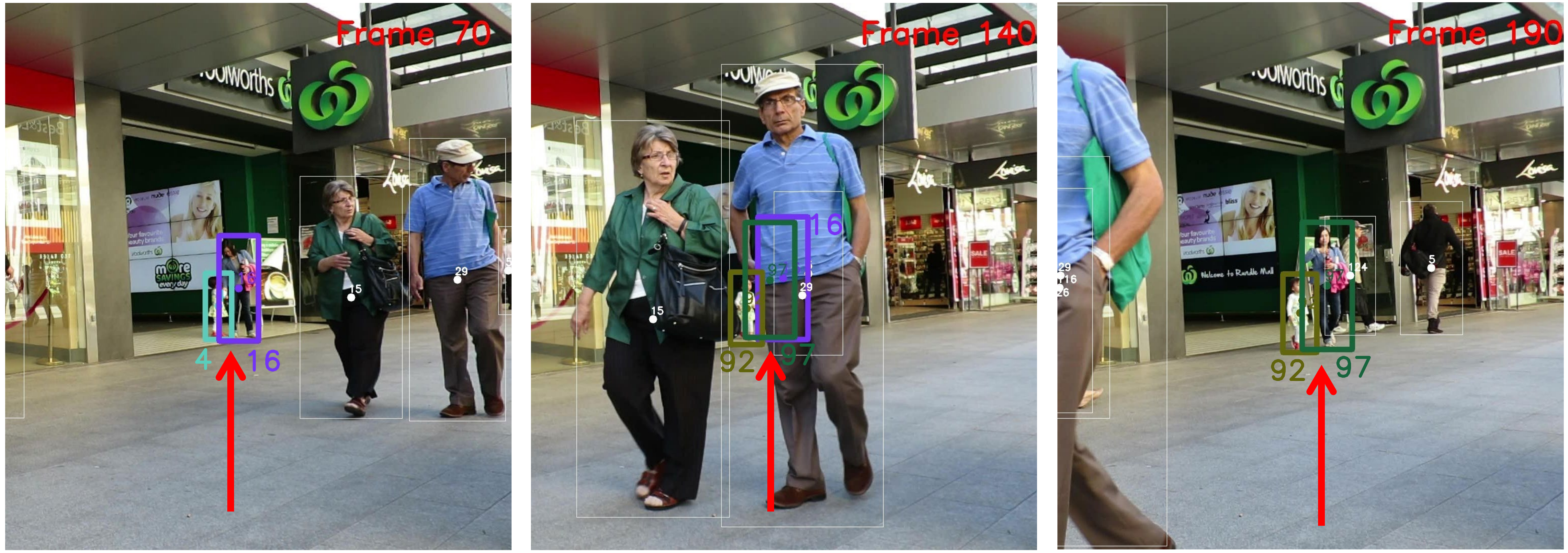}
        \raisebox{0.5in}{\rotatebox[origin=rc]{270}{MOT16-09}}
        
        \raisebox{0.5in}{\rotatebox[origin=lc]{90}{GLMB\_CSTrack}}
        \includegraphics[width=0.85\textwidth]{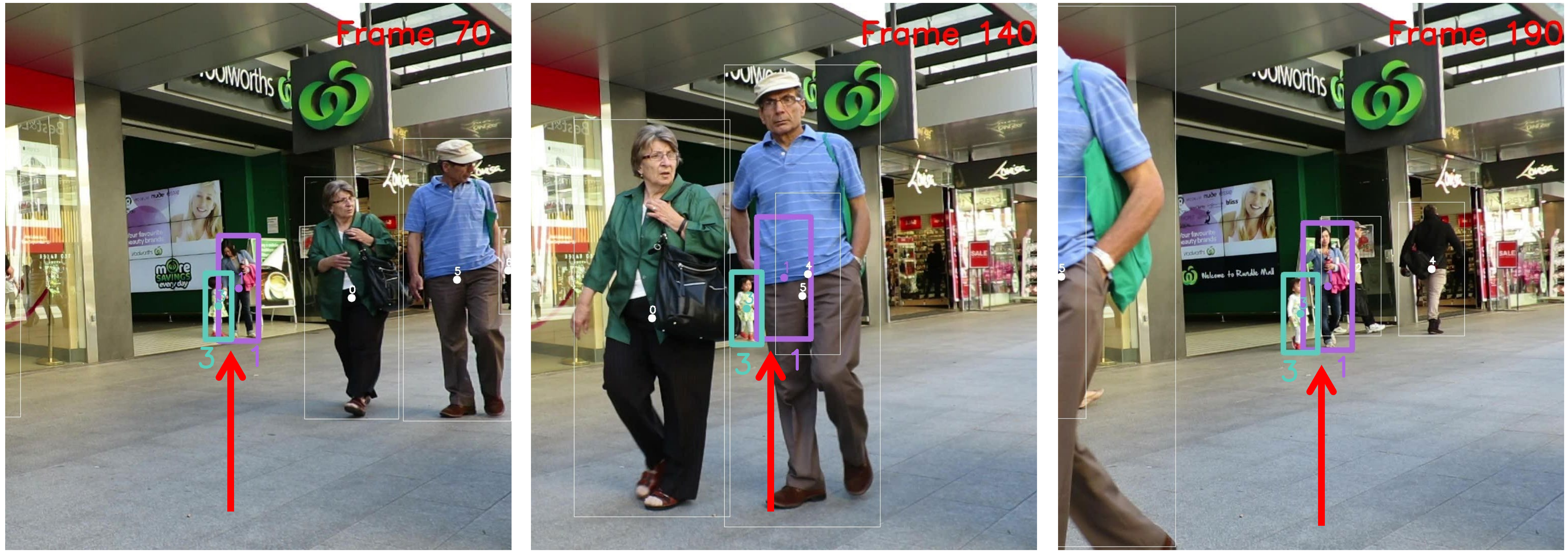}
        \raisebox{0.5in}{\rotatebox[origin=rc]{270}{MOT16-09}}

    \caption{Qualitative results of our GLMB filter compared with difference trackers. Numbers located in circles denote tracklet identity. For the sake of visualization, we only highlight tracklets that our method is different from others. The frame number is in the upper right of each image. We visually compare our results with ones from CSTrack. Best viewed in color and zoom.}

    \label{fig:quali_compa}
\end{figure*}


\begin{figure*}[h!]
\centering
  
        \raisebox{0.5in}{\rotatebox[origin=lc]{90}{GLMB\_CSTrack}}
        \includegraphics[width=0.85\textwidth]{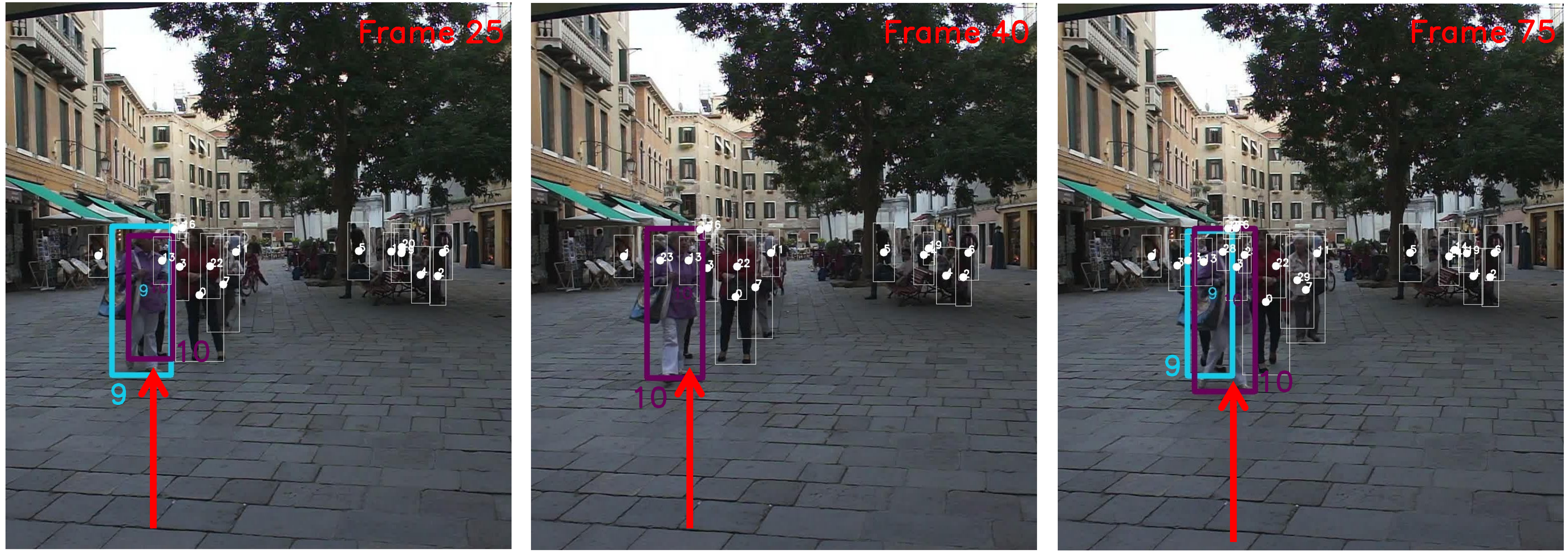}
        \raisebox{0.5in}{\rotatebox[origin=rc]{270}{MOT16-02}}\vspace{-0.57cm}

    \rule{0.925\textwidth}{.4pt}
        
        \raisebox{0.5in}{\rotatebox[origin=lc]{90}{GLMB\_CSTrack}}
        \includegraphics[width=0.85\textwidth]{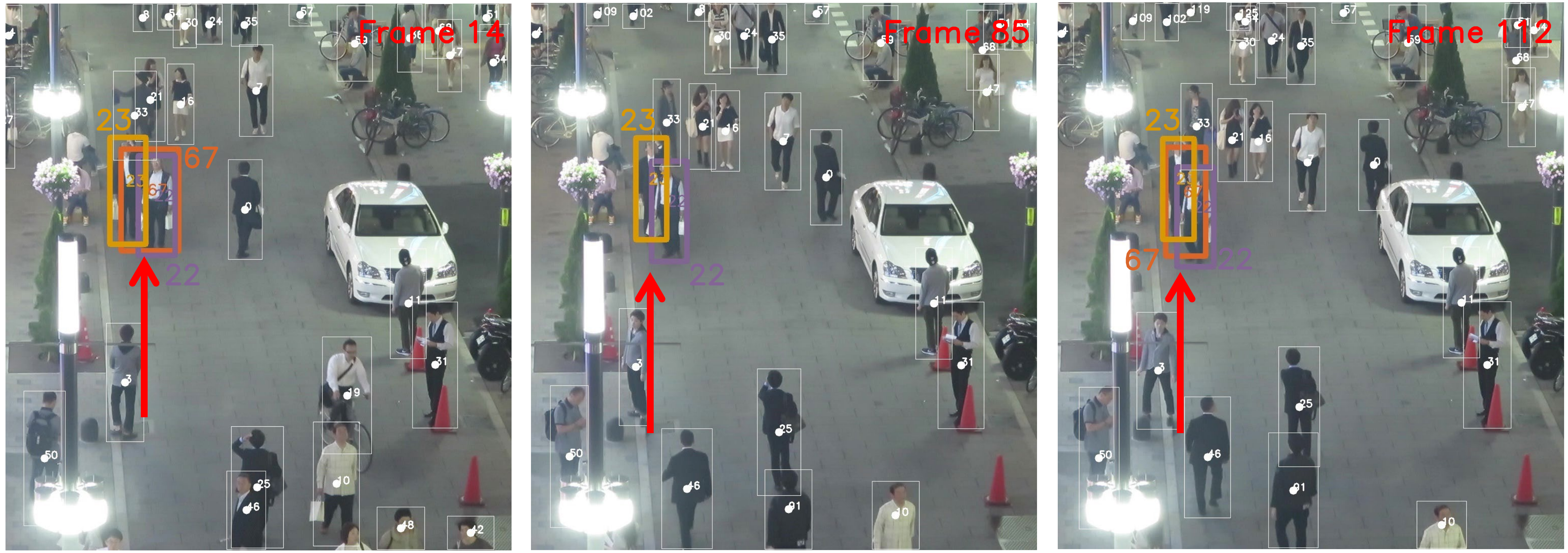}
        \raisebox{0.5in}{\rotatebox[origin=rc]{270}{MOT16-03}}
    
    \caption{Failure cases of GLMB tracker. We use CSTrack detection. Best viewed in color and zoom.}
    \label{fig:quali_compa_fail}
    
\end{figure*}


\subsubsection{Qualitative Comparison}
Figure~\ref{fig:quali_compa} shows our qualitative results in three sequences (i.e., MOT16-02 and MOT16-09) of MOT16 dataset. It shows that our methods are able to recall long-term occluded objects while CSTrack exhibits ID switches. For example, at frame 70 in MOT16-09 sequence, CSTrack assigns two ID 16 and 4 for objects annotated with purple and cyan bounding boxes, respectively. After occlusion, at frame 140, these 2 tracks are assigned new ID 97 and 92, respectively. In contrast, our tracker maintains the track ID consistently. Nevertheless, the proposed methods tend to mistake the track ID after occlusion when objects are visually similar. For example, in  Figure~\ref{fig:quali_compa_fail}, sequence MOT16-02, tracks 9 and 10 switch their ID after occlusion. Further, our methods occasionally initialize false tracks due to false positive detection. In sequence MOT16-03, there is a false positive detection generated between tracks 22 and 23, which creates the false positive track 67. Full videos related to the qualitative comparison are given in supplementary materials.

\subsection{Evaluation of Efficiency}\label{subsec:exp-efficiency}


\subsubsection{Run-Time Comparison}
Table~\ref{tbl:det_seq} presents the average run-time in frame per second (FPS) of all trackers in MOT16 validation dataset when the number of hypotheses $N_{h}$ is 500. The tests were performed on a desktop with CPU AMD Ryzen Threadripper 2950X 16-Core Processor (no GPU acceleration was used for our filters). Although our methods are slower than others, they can perform real-time tracking on MOT16 dataset with our hardware settings and implementations, given the highest frame rate sequence is 30 FPS in this dataset. 
\begin{table}[h!]
    \centering
    \caption{Run-time comparison with SOTA methods (\textcolor{red}{\textbf{red}}: the best, \textbf{bold}: the best in a detector, `$\ast$': our methods).} 
    \label{tbl:det_seq}
    \scriptsize
    \begin{tabular}{l l c|c l c}
    \toprule
        \textbf{Detector} & \textbf{Method} & \textbf{FPS}$\uparrow$ & \textbf{Detector} & \textbf{Method} & \textbf{FPS}$\uparrow$  \\ \midrule
        ~ & DeepSORT\cite{wojke2017simple} & \textbf{133.9}   & ~ & CSTrack & \textbf{250.1}   \\ 
        POI\cite{yu2016poi} & \textbf{LMB*} & 108.0   & CSTrack\cite{liang2020rethinking} & \textbf{LMB*} & 68.0   \\ 
        ~ & \textbf{GLMB*} & 104.0   & ~ & \textbf{GLMB*} &  58.3  \\ \midrule
        ~ & JDE & \textbf{101.4}   & ~ & FairMOT & \textbf{228.7}   \\ 
        JDE\cite{wang2020towards} & \textbf{LMB*} & 68.4   & FairMOT\cite{zhang2021fairmot} & \textbf{LMB*} & 87.7   \\ 
        ~ & \textbf{GLMB*} & 54.5   & ~ & \textbf{GLMB*} & 81.4   \\ \midrule
        ~ & TraDes & \textcolor{red}{\textbf{281.5}}   & ~ & GSDT & \textbf{266.7}   \\ 
        TraDes\cite{wu2021track} & \textbf{LMB*} & 111.5   & GSDT\cite{wang2020joint} & \textbf{LMB*} & 81.1   \\ 
        ~ & \textbf{GLMB*} & 96.2   & ~ & \textbf{GLMB*} & 72.0   \\ 
    \bottomrule
    \end{tabular}
\end{table}

\subsubsection{$N_{h}$ versus Tracking Accuracy}
\begin{figure*}[h!]
    \begin{centering}
        \includegraphics[width=\textwidth]{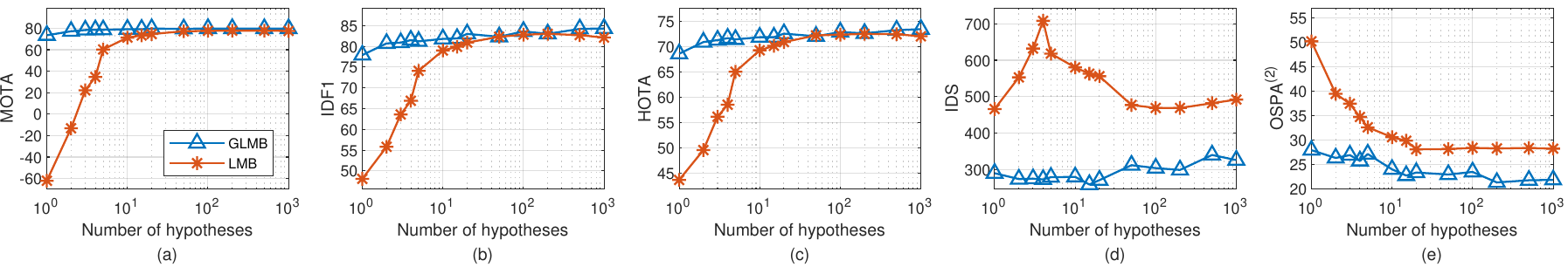} 
    \end{centering}
    \vspace{-1.0cm}
    \caption{Number of hypotheses ($N_{h}$) versus tracking accuracy for  GLMB and LMB filters in terms of MOTA, IDF1, HOTA, IDS scores and OSPA$^{\texttt{(2)}}$ distance (sub-figures (a) to (e), respectively). The numbers of hypotheses are plotted in log scale.}
    \label{fig:hyponum}
\end{figure*}

Figure~\ref{fig:hyponum} plots the relationship between $N_{h}$ and the accuracy of GLMB and LMB filters. We observe that when $N_{h}>50$, the accuracy of GLMB and LMB filters in terms of MOTA, IDF1, and HOTA is saturated while OSPA\!$^{\texttt{(2)}}$ error keep decreasing with higher $N_{h}$ for the GLMB filter. IDS for the GLMB filter is relatively consistent for different values of $N_{h}$. For the LMB filter, when the number of hypotheses is low (e.g., $N_h < 4$), the filter mostly estimates incorrect tracks. Since those tracks cannot be matched with the ground truth, IDS error is not captured (the error is mostly due to false positive and false negative). When $N_{h}$ increases (e.g., $N_h \geq 4$), more correct tracks are estimated. But for low $N_{h}$, there are not enough hypotheses to correctly resolve object ID, hence the increase in IDS error. When $N_{h}$ increases further, track ID is more consistent, hence the decrease in IDS error.


\subsection{Ablation Studies}\label{subsec:exp-ablation}
In this subsection, we perform a series of ablation studies for the proposed GLMB and LMB trackers on the MOT16 and MOT17 validation sets using FairMOT detector.
\subsubsection{Component-Wise Analysis}
In this experiment, we analyze the performance of our trackers under different configurations. The baseline (BL) setting (used in the previous experiments) includes the track appearance model, fuzzy detection model, and track re-ID module. To study the effects of different components, we test the following settings:
$\overline{\textrm{AR}}$ is the BL setting excluding object appearance model and track re-ID module;  $\overline{\textrm{R}}$ is the BL setting excluding the re-ID module; $\overline{\textrm{F}}$ is the BL setting excluding the fuzzy detection model (i.e., a constant detection probability is used).

The results in Table~\ref{tbl:ablfeat} show significant performance degradation in terms of HOTA, IDS and IDF1 for $\overline{\textrm{AR}}$ setting. Especially, for GLMB filter, OPSA\!$^{\texttt{(2)}}$ error increases significantly. A similar observation is for  $\overline{\textrm{R}}$ setting. It is because objects are assigned to new IDs after occlusion. This behavior highlights the importance of the proposed appearance model and track recall module in reducing ID switches. When the fuzzy detection model is removed, the drop in performance is not as significant as for the other settings. It can be explained because, with the fuzzy detection model, if the objects are not occluded and not too far from the camera, their detection probability is high (similar to the chosen constant detection probability). The detection probability decreases when the objects are occluded or far away from the camera to help maintain a reasonably high existence probability for these objects if they are actually miss-detected. Hence, the drop in performance for a constant detection probability model is not significant if the miss-detection rate is relatively low as that of the FairMOT detector.


\begin{table*}[t!]
\centering
\renewcommand{\arraystretch}{1}
\caption{Component-Wise Analysis for our methods (\textcolor{red}{\textbf{red}}: the best).}
\label{tbl:ablfeat}
    \scriptsize
     \begin{tabular}{p{0.1cm}|P{1.4cm}|p{1.6cm}|P{0.8cm} P{0.8cm} P{0.8cm} P{0.8cm} P{1cm}}
        \toprule
          \textbf{} & \textbf{Setting} & \textbf{Method} &   \textbf{MOTA$\uparrow$} & \textbf{IDF1$\uparrow$} &  \textbf{HOTA$\uparrow$} & \textbf{IDS$\downarrow$} & \textbf{OSPA\!$^{\texttt{(2)}}$\!$\downarrow$}\\
        \midrule
            \multirow{7}{*}{\rotatebox{90}{\textbf{MOT16}}}
            & \multirow{2}{*}{BL} & \textbf{LMB} &  \textcolor{red}{\textbf{81.9}}  & 82.1  & \textcolor{red}{\textbf{67.3}}  & 503  & \textcolor{red}{\textbf{26.2}} \\
            & & \textbf{GLMB}    & 81.3 & \textcolor{red}{\textbf{83.1}} & 67.2  & \textcolor{red}{\textbf{439}} & 27.3\\
            \cline{2-8}
            & \multirow{2}{*}{$\overline{\textrm{AR}}$} & \textbf{LMB} & 80.0  & 74.6  & 62.8  & 808  & 36.2 \\
            & & \textbf{GLMB} & 78.6  & 71.7  & 61.3  & 1,254  & 60.7\\
            \cline{2-8}
            & \multirow{2}{*}{$\overline{\textrm{R}}$} & \textbf{LMB} & 81.1  & 79.2  & 66.0  & 637  & 46.9 \\
            & & \textbf{GLMB} & 80.6  & 76.2  & 64.1  & 837  & 55.4\\
            \cline{2-8}
            & \multirow{2}{*}{$\overline{\textrm{F}}$} & \textbf{LMB} & 78.9  & 80.0  & 65.5  & 486  & 30.43 \\
            & & \textbf{GLMB} & 79.5 & 79.7  & 65.4  & 633 & 31.48\\
        \hline
            \multirow{8}{*}{\rotatebox{90}{\textbf{MOT17}}}
            & \multirow{2}{*}{BL} & \textbf{LMB} &  \textcolor{red}{\textbf{82.3}}  & 82.2  & \textcolor{red}{\textbf{67.3}}  & 1,566  & \textcolor{red}{\textbf{24.3}} \\
            & & \textbf{GLMB} & 81.9 & \textcolor{red}{\textbf{82.9}} & \textcolor{red}{\textbf{67.3}} & \textcolor{red}{\textbf{1,512}} & 25.5\\
            \cline{2-8}
            & \multirow{2}{*}{$\overline{\textrm{AR}}$} & \textbf{LMB} & 80.3  & 74.4  & 62.6  & 2,532  & 33.0\\
            & & \textbf{GLMB} & 78.5  & 71.4  & 61.0  & 3,960  & 58.5\\
            \cline{2-8}
            & \multirow{2}{*}{$\overline{\textrm{R}}$} & \textbf{LMB} & 81.7  & 79.2  & 66.0  & 1,986  & 43.9 \\
            & & \textbf{GLMB} & 80.6 & 75.9  & 64.0  & 2,619 & 53.0\\
            \cline{2-8}
            & \multirow{2}{*}{$\overline{\textrm{F}}$} & \textbf{LMB} & 78.7  & 79.8  & 65.3  & 1,512  & 27.5 \\
            & & \textbf{GLMB} & 79.3 & 79.5  & 65.2  & 2,007 & 28.1\\
        \bottomrule
    \end{tabular}
\end{table*}

\subsubsection{Recall Length Analysis}
\begin{figure*}[h!]
    \begin{centering}
    \includegraphics[width=\textwidth]{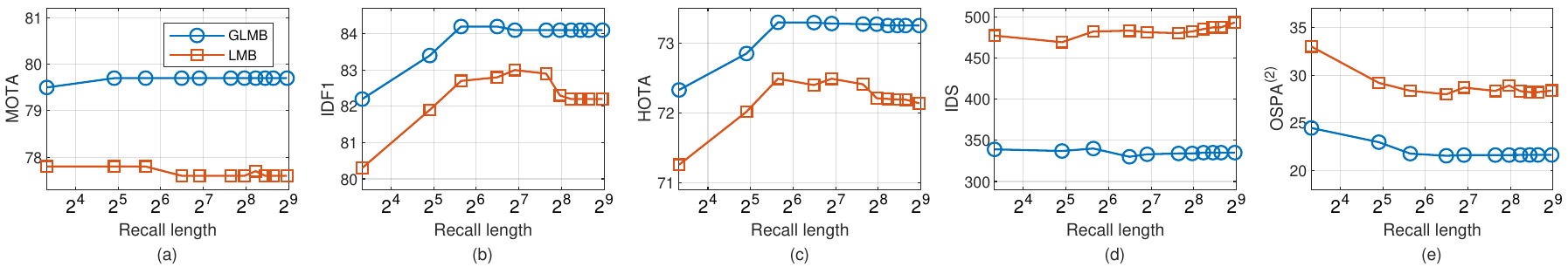} 
    \par\end{centering}
    \caption{Recall length versus tracking accuracy for GLMB and LMB filters in terms of MOTA, IDF1, HOTA, IDS scores and OSPA$^{\texttt{(2)}}$ distance (sub-figures (a) to (e), respectively). The recall lengths are plotted in base-two log scale.\label{fig:recall_len}}
\end{figure*}

The number of frames that we store a TT track before it is recalled (i.e., the recall length) affects the filters' performance. This experiment studies the tracking accuracy of the filters for various recall lengths. Figure~\ref{fig:recall_len} demonstrates that for the GLMB filter, the performance increases and then saturates when the number of recall frames is up to 50. Nevertheless, for the LMB filter, the performance drops slightly when the number of recall frames is beyond 200. The saturation in performance is either due to the fact that most tracks reappear after 50 frames or the appearance feature vectors of the tracks (extracted by the detector) change significantly after 50 frames. The drop in LMB filter performance for a high number of recall frames is due to the increase in the number of tracks that the filter needs to handle (higher possibility of track recall when recall length increases), hence degradation in data association (note that data association quality in LMB filter is lower than of GLMB filter due to the LMB approximation). 

\subsection{Fuzzy Rules Analysis}\label{subsec:fuzzyabl}
In this ablation study, we analyze the effects of the fuzzy rules on tracking performance. In Table \ref{tab:fuzzyabl}, we list the tested rules, in which rules R1, R2, R3 (R1 is the baseline rule presented in Table \ref{tab:fuzzy-Relationship-between-membership}) follow the model intuition discussed in Subsection \ref{subsec:fuzzy} (i.e., the trend of IoA score contradicts the trend of the detection
probability, and the trend of area ratio follows the trend of the detection probability), while rules R4 and R5 contradict the model intuition. The results in Table \ref{tab:fuzzyabl_result} show that rules that follow the model intuition show similar tracking performance to the baseline. In contrast, we observe the degradation in performance if the rule is counter-intuitive (could be worse than not having the fuzzy model, compared to the $\bar{F}$ rows in Table \ref{tbl:ablfeat}).

\begin{table*}[h!]
\scriptsize
\renewcommand{\arraystretch}{1}
\caption{Tested fuzzy rules. R1, R2 and R3 follow the model intuition. R4 and R5 contradict the model intuition. \label{tab:fuzzyabl}}
\centering
    \begin{tabular}{c|ccc|ccc|ccc|ccc|ccc}
    \toprule 
    \textbf{Rules} & \multicolumn{3}{c|}{\textbf{R1}} & \multicolumn{3}{c|}{\textbf{R2}} & \multicolumn{3}{c|}{\textbf{R3}} & \multicolumn{3}{c|}{\textbf{R4}} & \multicolumn{3}{c}{\textbf{R5}}\tabularnewline
    \hline 
    \diagbox[height=0.6cm]{\hspace{-0.25cm} \textbf{Area ratio}}{\textbf{IoA}} & \multicolumn{1}{c}{\textbf{L}} & \multicolumn{1}{c}{\textbf{M}} & \multicolumn{1}{c|}{\textbf{H}} & \multicolumn{1}{c}{\textbf{L}} & \multicolumn{1}{c}{\textbf{M}} & \multicolumn{1}{c|}{\textbf{H}} & \multicolumn{1}{c}{\textbf{L}} & \multicolumn{1}{c}{\textbf{M}} & \multicolumn{1}{c|}{\textbf{H}} & \multicolumn{1}{c}{\textbf{L}} & \multicolumn{1}{c}{\textbf{M}} & \multicolumn{1}{c|}{\textbf{H}} & \multicolumn{1}{c}{\textbf{L}} & \multicolumn{1}{c}{\textbf{M}} & \multicolumn{1}{c}{\textbf{H}}\tabularnewline
    \midrule 
    \textbf{L} & M & L & L & M & L & L & M & M & L & L & H & H & M & H & M\tabularnewline
    \textbf{M} & M & M & L & H & M & L & M & M & L & L & L & H & L & L & H\tabularnewline
    \textbf{H} & H & H & L & H & H & M & H & H & M & L & L & H & L & L & H\tabularnewline
    \bottomrule
    \end{tabular}
\end{table*}

\begin{table}[h!]
\renewcommand{\arraystretch}{1}
\caption{Our evaluation result with FairMOT detector on five different fuzzy rules. \label{tab:fuzzyabl_result}}
\centering
\scriptsize
\begin{tabular}{p{0.1cm}|p{0.8cm}|p{0.7cm}|p{0.6cm}p{0.6cm}p{0.6cm}p{0.6cm}p{0.8cm}}
\toprule 
 & \textbf{Method} & \textbf{Setting}  &   \textbf{MOTA$\uparrow$} & \textbf{IDF1$\uparrow$} &  \textbf{HOTA$\uparrow$} & \textbf{IDS$\downarrow$} & \textbf{OSPA\!$^{\texttt{(2)}}$\!$\downarrow$} \tabularnewline
\midrule 
\multirow{10}{*}{\rotatebox{90}{\textbf{MOT16}}} & \multirow{5}{*}{LMB} & R1 & \textbf{\textcolor{red}{81.9}} & 82.1 & 67.3 & 503 & \textbf{\textcolor{red}{26.3}}\tabularnewline
 &  & R2 & 80.8 & 82.3 & 67.2 & 482 & 27.1\tabularnewline
 &  & R3 & 80.9 & \textbf{\textcolor{red}{82.4}} & \textbf{\textcolor{red}{67.3}} & \textbf{\textcolor{red}{476}} & 26.8\tabularnewline
 &  & R4 & 75.9 & 79.8 & 64.6 & 490 & 29.2\tabularnewline
 &  & R5 & 77.6 & 80.6 & 65.5 & 481 & 27.5\tabularnewline
\cline{2-8} \cline{3-8} \cline{4-8} \cline{5-8} \cline{6-8} \cline{7-8} \cline{8-8} 
 & \multirow{5}{*}{GLMB} & R1 & \textcolor{red}{\textbf{81.3}} & \textbf{\textcolor{red}{83.1}} & \textbf{\textcolor{red}{67.2}} & 439 & \textbf{\textcolor{red}{27.3}}\tabularnewline
 &  & R2 & \textbf{\textcolor{red}{81.3}} & 82.1 & 66.8 & \textbf{\textcolor{red}{438}} & 28.6\tabularnewline
 &  & R3 & 81.2 & 82.4 & 66.9 & 485 & 28.1\tabularnewline
 &  & R4 & 78.1 & 79.0 & 64.6 & 521 & 32.0\tabularnewline
 &  & R5 & 77.9 & 78.5 & 64.3 & 473 & 30.5\tabularnewline
\hline 
\multirow{10}{*}{\rotatebox{90}{\textbf{MOT17}}} & \multirow{5}{*}{LMB} & R1 & \textbf{\textcolor{red}{82.3}} & 82.2 & \textbf{\textcolor{red}{67.3}} & 1,566 & \textbf{\textcolor{red}{24.3}}\tabularnewline
 &  & R2 & 81.3 & 82.3 & 67.2 & 1,500 & 25.2\tabularnewline
 &  & R3 & 81.4 & \textbf{\textcolor{red}{82.4}} & \textbf{\textcolor{red}{67.3}} & \textbf{\textcolor{red}{1,479}} & 24.9\tabularnewline
 &  & R4 & 76.4 & 79.9 & 64.6 & 1,491 & 27.3\tabularnewline
 &  & R5 & 78.0 & 80.7 & 65.5 & 1,482 & 25.7\tabularnewline
\cline{2-8} \cline{3-8} \cline{4-8} \cline{5-8} \cline{6-8} \cline{7-8} \cline{8-8} 
 & \multirow{5}{*}{GLMB} & R1 & \textbf{\textcolor{red}{81.9}} & \textcolor{red}{\textbf{82.9}} & \textbf{\textcolor{red}{67.3}} & 1,512 & \textbf{\textcolor{red}{25.5}}\tabularnewline
 &  & R2 & 81.6 & 82.1 & 66.8 & \textbf{\textcolor{red}{1,401}} & 26.8\tabularnewline
 &  & R3 & 81.6 & 82.5 & 66.9 & 1,548 & 26.3\tabularnewline
 &  & R4 & 78.5 & 79.0 & 64.6 & 1,641 & 29.7\tabularnewline
 &  & R5 & 78.4 & 78.6 & 64.3 & 1,527 & 28.7\tabularnewline
\hline 
\end{tabular}
\end{table}


\subsection{Limitation}\label{subsec:limitation}

Considering multiple hypotheses of the data association comes with high computational cost, although Gibbs sampling can sample significant data association hypotheses at the linear complexity in the number of measurements \cite{vo2016efficient} (in fact standard rank assignment solver such as Murty algorithm may become unusable for a large dataset such as MOT20). Hence, our trackers are generally slower than SOTA trackers, especially, in dataset involving large number of objects. However, keeping multiple hypotheses contributes less number of ID switches when in the crowded scenario for MOT20 test set as can be seen in Table~\ref{tbl:testset}. Conversely, aspect ratios of the bounding boxes and object heights are modeled with Gaussian distributions in our work. However, since they can only be positive, constraints need to be imposed in the state estimation step to ensure the corresponding estimated values are positive.

\section{Conclusion}\label{sec:conclusion}
Using an RFS (simple finite point process) formulation, we have developed multi-object Bayes filters that address occlusion and re-ID in visual MOT. The proposed filters generally outperform benchmarking SOTA methods on MOTchallenge datasets. Further, they exhibit a much lower number of ID switches compared to SOTA methods, thus validating the effectiveness of the proposed filter’s occlusion resolution and re-ID capability. This is mainly due to the processing of multiple data association hypotheses rather than a single hypothesis as in the widely used benchmarking algorithms. The trade-off is a slower processing time, nonetheless, it is more than enough for real-time processing even with a prototype implementation. From an analytical viewpoint, the proposed filter scales gracefully with problem size, with linear complexity in the number of detections. 

In terms of future work, the current measurement model only uses bounding box measurements from the detector, excluding the detection confidence information. Nevertheless, it has been shown that trackers that can handle the detection confidence score improve the tracking performance \cite{zhang2022bytetrack}. Thus, an extension of this work could be the development of a measurement model that can handle detection confidence scores. Moreover, the object dynamic model in this work can also be extended to include complex object poses \cite{dang2024kinematics}. The incorporation of the pose estimation model, on the one hand, could improve tracking performance if the poses can be modeled accurately. On the other hand, it increases the applicability of the trackers to a wider range of applications. Nevertheless, these extensions come at a high computational expense, which could decrease the efficiency of the trackers.

\bibliographystyle{IEEEtran}
\bibliography{reflibformated.bib}

\clearpage
\newpage 
\setcounter{section}{0}
\setcounter{equation}{0}

\twocolumn[
\begin{@twocolumnfalse}
\vspace*{1in}
\begin{center}
    \LARGE\bf Supplementary Materials: Visual Multi-Object Tracking with Object Appearance-Reappearance Resolution and Occlusion Handling using Labeled Random Finite Set
\end{center}
\vspace*{0.5in}
\vspace*{0.5in}
\end{@twocolumnfalse}
]

\section{Details on Filtering Recursions}\label{sec:appendix}

\subsection{Exact Filtering Recursion}\label{subsec:app-exact}

Given the prior multi-object density $\boldsymbol{\pi}$
with the form of a general multi-object density, applying the dynamic and measurement models allows us to
derive the (unnormalized) filtering multi-object density
\allowdisplaybreaks
\begin{multline}
\Omega\left(\boldsymbol{\pi},\{P_{B,+}^{(\ell)},f_{B,+}^{(\ell)}\}_{\ell\in\mathbb{B}_{+}},Z_{+}\right)\propto\Delta(\boldsymbol{X}_{+}) \\
\sum_{I,\xi,I_{+},\theta_{+}}\delta_{I_{+}}[\mathcal{L}(\boldsymbol{X}_{+})]\omega_{Z_{+},\boldsymbol{X}_{+}}^{(I,\xi,I_{+},\theta_{+})}[p_{Z_{+},\boldsymbol{X}_{+}}^{(\xi,\theta_{+})}(\cdot)]^{\boldsymbol{X}_{+}},\label{eq:exact_GLMB_filtering}
\end{multline} 
where 
\begin{eqnarray*}
\omega_{Z,\boldsymbol{X}}^{(I,\xi,I_{+},\theta_{+})} &\!\!\!\!\! \propto \!\!\!\!\! & \omega^{(I,\xi)}[\bar{P}_{S}^{(\xi)}]^{I\cap I_{+}}[1-\bar{P}_{S}^{(\xi)}]^{I-I_{+}} \\ 
& \!\!\!\!\!\!\!\!\!\! & [P_{B}]^{I_{+}\cap\mathbb{B}_{+}}[1-P_{B}]^{\mathbb{B}_{+}-I_{+}}[\bar{\psi}_{Z,\boldsymbol{X}}^{(\xi,\theta_{+})}]^{I_{+}},\\
\bar{P}_{S}^{(\xi)}(\ell) & \!\!\!\!\!=\!\!\!\!\! & \langle p^{(\xi)}(\cdot,\ell),P_{S}(\cdot,\ell)\rangle,\\
\bar{\psi}_{Z,\boldsymbol{X}}^{(\xi,\theta)}(\ell) &\!\!\!\!\! =\!\!\!\!\! & \langle p_{+}^{(\xi)}(\cdot,\ell),\psi_{Z,\boldsymbol{X}}^{(\theta)}(\cdot,\ell)\rangle,\\
p_{+}^{(\xi)}(x,\ell) & \!\!\!\!\!=\!\!\!\!\! & 1_{\mathbb{L}}(\ell)\langle P_{S}(\cdot,\ell)p^{(\xi)}(\cdot,\ell),f_{S,+}^{(\ell)}(x|\cdot)\rangle/\bar{P}_{S}^{(\xi)}(\ell)\\
& \!\!\!\!\!\!\!\!\!\! & +1_{\mathbb{B}_{+}}(\ell)f_{B,+}^{(\ell)}(x),\\
p_{Z,\boldsymbol{X}}^{(\xi,\theta)}(x,\ell) & \!\!\!\!\!=\!\!\!\!\! & p_{+}^{(\xi)}(x,\ell)\psi_{Z,\boldsymbol{X}}^{(\theta)}(x,\ell)/\bar{\psi}_{Z,\boldsymbol{X}}^{(\xi,\theta_{+})}(\ell).
\end{eqnarray*}

Note that the above filtering density is not a GLMB density since
each term of the product $[p_{Z,\boldsymbol{X}_{+}\backslash\{\cdot\}}^{(\xi,\theta_{+})}(\cdot)]^{\boldsymbol{X}_{+}}$
depends on $\boldsymbol{X}_{+}$. Hence, $[p_{Z,\boldsymbol{X}_{+}}^{(\xi,\theta_{+})}(\cdot)]^{\boldsymbol{X}_{+}}$
can be effectively written as $p_{Z_{+}}^{(\xi,\theta_{+})}(\boldsymbol{X}_{+})$,
from which the filtering density (\ref{eq:exact_GLMB_filtering})
indeed takes the form
\begin{multline*}
\Omega\left(\boldsymbol{\pi},\{P_{B,+}^{(\ell)},f_{B,+}^{(\ell)}\}_{\ell\in\mathbb{B}_{+}},Z_{+}\right)\propto\Delta(\boldsymbol{X}_{+})\\
\!\!\!\sum_{I,\xi,I_{+},\theta_{+}}\!\!\!\delta_{I_{+}}[\mathcal{L}(\boldsymbol{X}_{+})]\omega_{Z_{+},\boldsymbol{X}_{+}}^{(I,\xi,I_{+},\theta_{+})}p_{Z_{+}}^{(\xi,\theta_{+})}(\boldsymbol{X}_{+}),
\end{multline*}
which is not a GLMB density.

\subsection{GLMB Filtering Recursion} \label{subsec:app-glmb}

Given the prior multi-object density $\boldsymbol{\pi}$
with the form of a GLMB density, applying the
dynamic and measurement models allows us to
derive the (unnormalized) filtering multi-object density (by following
Proposition 1 of \cite{vo2016efficient})
\allowdisplaybreaks
\begin{multline}
\Omega\left(\boldsymbol{\pi},\{P_{B,+}^{(\ell)},f_{B,+}^{(\ell)}\}_{\ell\in\mathbb{B}_{+}},Z_{+}\right)\propto\Delta(\boldsymbol{X}_{+}) \\
\!\!\!\sum_{I,\xi,I_{+},\theta_{+}}\!\delta_{I_{+}}[\mathcal{L}(\boldsymbol{X}_{+})]\omega_{Z_{+},\hat{\boldsymbol{X}}_{+}^{(\xi,I_{+})}}^{(I,\xi,I_{+},\theta_{+})}[p_{Z_{+},\hat{\boldsymbol{X}}_{+}^{(\xi,I_{+})}}^{(\xi,\theta_{+})}(\cdot)]^{\boldsymbol{X}_{+}},\label{eq:GLMB_filtering_app}
\end{multline}
where $\hat{\boldsymbol{X}}_{+}^{(\xi,I_{+})}=\{(\hat{x}_{+}^{(\xi,\ell)},\ell):\ell\in I_{+}\}$, $\hat{x}_{+}^{(\xi,\ell)}$ is the estimate from the prediction density $p_{+}^{(\xi)}(\cdot,\ell)$, and
\begin{eqnarray*}
\omega_{Z,\hat{\boldsymbol{X}}}^{(I,\xi,I_{+},\theta_{+})} &\!\!\!\!\! \propto \!\!\!\!\! & \omega^{(I,\xi)}[\bar{P}_{S}^{(\xi)}]^{I\cap I_{+}}[1-\bar{P}_{S}^{(\xi)}]^{I-I_{+}} \\ 
& \!\!\!\!\!\!\!\!\!\! & [P_{B}]^{I_{+}\cap\mathbb{B}_{+}}[1-P_{B}]^{\mathbb{B}_{+}-I_{+}}[\bar{\psi}_{Z,\hat{\boldsymbol{X}}}^{(\xi,\theta_{+})}]^{I_{+}},\\
\bar{P}_{S}^{(\xi)}(\ell) & \!\!\!\!\!=\!\!\!\!\! & \langle p^{(\xi)}(\cdot,\ell),P_{S}(\cdot,\ell)\rangle,\\
\bar{\psi}_{Z,\hat{\boldsymbol{X}}}^{(\xi,\theta)}(\ell) &\!\!\!\!\! =\!\!\!\!\! & \langle p_{+}^{(\xi)}(\cdot,\ell),\psi_{Z,\hat{\boldsymbol{X}}}^{(\theta)}(\cdot,\ell)\rangle,\\
p_{+}^{(\xi)}(x,\ell) & \!\!\!\!\!=\!\!\!\!\! & 1_{\mathbb{L}}(\ell)\langle P_{S}(\cdot,\ell)p^{(\xi)}(\cdot,\ell),f_{S,+}^{(\ell)}(x|\cdot)\rangle/\bar{P}_{S}^{(\xi)}(\ell)\\
& \!\!\!\!\!\!\!\!\!\! & +1_{\mathbb{B}_{+}}(\ell)f_{B,+}^{(\ell)}(x),\\
p_{Z,\hat{\boldsymbol{X}}}^{(\xi,\theta)}(x,\ell) & \!\!\!\!\!=\!\!\!\!\! & p_{+}^{(\xi)}(x,\ell)\psi_{Z,\hat{\boldsymbol{X}}}^{(\theta)}(x,\ell)/\bar{\psi}_{Z,\hat{\boldsymbol{X}}}^{(\xi,\theta_{+})}(\ell).
\end{eqnarray*}

\subsection{LMB Filtering Recursion}\label{subsec:app-lmb}
With the same analogy, if the prior density is an LMB density, we have a special case:
\begin{multline}
\Omega\left(\boldsymbol{\pi},\{P_{B,+}^{(\ell)},f_{B,+}^{(\ell)}\}_{\ell\in\mathbb{B}_{+}},Z_{+}\right) \propto \Delta(\boldsymbol{X}_{+})\\
\sum_{I_{+},\theta_{+}}\delta_{I_{+}}[\mathcal{L}(\boldsymbol{X}_{+})]\omega_{Z_{+},\hat{\boldsymbol{X}}_{+}^{(I_{+})}}^{(I_{+},\theta_{+})}[p_{Z_{+},\hat{\boldsymbol{X}}_{+}^{(I_{+})}}^{(\theta_{+})}(\cdot)]^{\boldsymbol{X}_{+}},\label{eq:exact_GLMB_LMB_filtering_app}
\end{multline}
where $\hat{\boldsymbol{X}}_{+}^{(I_{+})}=\{(\hat{x}_{+}^{(\ell)},\ell):\ell\in I_{+}\}$ with $\hat{x}_{+}^{(\ell)}$ is the estimate from the
prediction density $p_{+}(\cdot,\ell)$,
\begin{eqnarray*}
    \omega_{Z,\hat{\boldsymbol{X}}}^{(I_{+},\theta_{+})} &\!\!\!\!\! \propto\!\!\!\!\! & \sum_{(I_{+}-\mathbb{B}_{+})\supseteq I}\!\!\!\!w(I)[\bar{P}_{S}]^{I\cap I_{+}}[1-\bar{P}_{S}]^{I-I_{+}} \\ &\!\!\!\!\!\!\!\!\!\!&[P_{B}]^{I_{+}\cap\mathbb{B}_{+}}[1-P_{B}]^{\mathbb{B}_{+}-I_{+}}[\bar{\psi}_{Z,\hat{\boldsymbol{X}}}^{(\theta_{+})}]^{I_{+}},\\
    \bar{P}_{S}(\ell) &\!\!\!\!\! = \!\!\!\!\!& \langle p(\cdot,\ell),P_{S}(\cdot,\ell)\rangle,\\
    \bar{\psi}_{Z,\hat{\boldsymbol{X}}}^{(\theta)}(\ell) &\!\!\!\!\! = \!\!\!\!\!& \langle p_{+}(\cdot,\ell),\psi_{Z,\hat{\boldsymbol{X}}}^{(\theta)}(\cdot,\ell)\rangle,\\
    p_{+}(x,\ell) &\!\!\!\!\!= \!\!\!\!\!& 1_{\mathbb{L}}(\ell)\langle P_{S}(\cdot,\ell)p(\cdot,\ell),f_{S,+}^{(\ell)}(x|\cdot)\rangle/\bar{P}_{S}(\ell)\\ &\!\!\!\!\!\!\!\!\!\!&+1_{\mathbb{B}_{+}}(\ell)f_{B,+}^{(\ell)}(x),\\
     p_{Z,\hat{\boldsymbol{X}}}^{(\theta)}(x,\ell) & \!\!\!\!\!=\!\!\!\!\! & p_{+}(x,\ell)\psi_{Z,\hat{\boldsymbol{X}}}^{(\theta)}(x,\ell)/\bar{\psi}_{Z,\hat{\boldsymbol{X}}}^{(\theta)}(\ell).
\end{eqnarray*}

\section{OSPA\protect\textsuperscript{(2)} Metric}\label{subsec:app-ospa}
Consider a metric space $(\mathcal{\mathbb{W}},\underline{d})$ with
the \emph{base-distance} $\underline{d}:\mathcal{\mathcal{\mathbb{W}}\times}\mathcal{\mathbb{W}}\rightarrow[0;\infty)$
between the elements of $\mathcal{\mathbb{W}}$ (a set of all bounding
boxes in our context), the general form of OSPA distance of order
$p\geq1$, and cut-off $c>0$, between two sets $X=\{x_{1},...,x_{m}\}$
and $Y=\{y_{1},...,y_{n}\}$ is defined by \cite{schuhmacher2008aconsistent,rezatofighi2020trustworthy} 
\begin{multline}
    d_{\mathtt{O}}^{(p,c)}(X,Y)=\\
\left(\frac{1}{n}\left(\min_{\pi\in\Pi_{n}}\sum_{i=1}^{m}\underline{d}^{(c)}\left(x_{i},y_{\pi(i)}\right)^{p}+c^{p}\left(n-m\right)\right)\right)^{\frac{1}{p}},\label{eq:OSPA-dist-c}
\end{multline}
if $n\geq m>0$, where $\Pi_{n}$ is the set of permutations of $\left\{ 1,2,...,n\right\} $
and $\underline{d}^{(c)}(x,y)=\min\left(c,\underline{d}\left(x,y\right)\right)$.
If one of the set is empty $d_{\mathtt{O}}^{(p,c)}(X,Y)=c$ , and
if both sets are empty $d_{\mathtt{O}}^{(p,c)}(\emptyset,\emptyset)=0$.

In a discrete-time window $\mathbb{T}$, a \textit{track} in a metric
space $(\mathcal{\mathbb{W}},\underline{d})$ can be defined as a
mapping $f:\mathbb{T}\mapsto\mathbb{W}$ \cite{beard2020asolution}. Its \textit{domain}
$\mathcal{D}_{f}\subseteq\mathbb{T}$, is the set of time instants
when the object/track has a state in $\mathbb{W}$. The general OSPA
distance above yields the following meaningful base-distance between
two tracks $f$ and $g$: 
\begin{align*}
\underline{d}^{\left(c\right)}\left(f,g\right)= & \sum\limits _{t\in\mathcal{D}_{f}\cup\mathcal{D}_{g}}\!\frac{d_{\mathtt{O}}^{\left(c\right)}\left(\left\{ f\left(t\right)\right\} ,\left\{ g\left(t\right)\right\} \right)}{\left|\mathcal{D}_{f}\cup\mathcal{D}_{g}\right|},
\end{align*}
if $\mathcal{D}_{f}\cup\mathcal{D}_{g}\neq\emptyset$, and $\underline{d}^{\left(c\right)}\left(f,g\right)=0$
if $\mathcal{D}_{f}\cup\mathcal{D}_{g}=\emptyset$, where $d_{\mathtt{O}}^{\left(c\right)}$
denotes the OSPA distance (the order parameter $p$ is redundant because
only sets of at most one element are considered) \cite{beard2020asolution}.

The task of evaluating the tracking results can be cast as measuring the
distance between two sets of tracks. This distance can be measured
with OSPA metric (consider $X$ and $Y$ in (\ref{eq:OSPA-dist-c})
as two sets of tracks) with $\underline{d}^{\left(c\right)}$ as the
base-distance. Since OSPA metric is applied twice (once to measure
the distance between two tracks and once to measure the distance between
two sets of tracks), we have the name OSPA\textsuperscript{(2)} metric
\cite{beard2020asolution}. In this work, we use the Generalized Intersection over Union (GIoU)
distance (i.e., $1-GIoU$, where the $GIoU$ score between bounding boxes are
computed according to \cite{rezatofighi2019giou}) as our base-distance in $\mathcal{\mathbb{W}}$. Since GIoU
base-distance is bounded by 1, we set $c=1$ to be sensitive to the
whole range of localization error.
whole range of localization error.

\end{document}